\documentclass[journal]{IEEEtran}

\newtheorem{definition}{Definition}

\usepackage{multirow}
\usepackage{amsmath}
\usepackage{amssymb}
\usepackage{tabularx}
\usepackage{booktabs} 
\usepackage{adjustbox}
\usepackage[table]{xcolor}
\usepackage{graphicx}
\usepackage{makecell}

\usepackage{xcolor}
\usepackage{colortbl}
\definecolor{light-gray}{gray}{0.90}

\usepackage{algpseudocode}
\usepackage[ruled,linesnumbered]{algorithm2e}
\usepackage[switch]{lineno}  
\usepackage{amsmath, amssymb}

\ifCLASSOPTIONcompsoc
  \usepackage[nocompress]{cite}
\else
  \usepackage{cite}
\fi

\usepackage{array}

  \usepackage[caption=false,font=footnotesize,labelfont=sf,textfont=sf]{subfig}

  \usepackage[caption=false,font=footnotesize]{subfig}

\newcommand\MYhyperrefoptions{bookmarks=true,bookmarksnumbered=true,
	pdfpagemode={UseOutlines},plainpages=false,pdfpagelabels=true,
	colorlinks=true,linkcolor={blue},citecolor={blue},urlcolor={blue},
	pdftitle={GraphIFE: Rethinking Graph Imbalance Node Classification From Invariant Learning},
	pdfsubject={Typesetting},
	pdfauthor={Fanlong Zeng},
	pdfkeywords={Graph, imbalance learning, node classification, data augmentation.}}
\ifCLASSINFOpdf
\usepackage[\MYhyperrefoptions,pdftex]{hyperref}
\else
\usepackage[\MYhyperrefoptions,breaklinks=true,dvips]{hyperref}
\usepackage{breakurl}
\fi

\usepackage{url}

\begin{document}

\title{GraphIFE: Rethinking Graph Imbalance Node Classification via Invariant Learning}

\author{Fanlong Zeng, Wensheng Gan, Kangjie Chen, Philip S. Yu,~\IEEEmembership{Life Fellow,~IEEE}
	
\thanks{Corresponding author: Wensheng Gan}
		
\thanks{Fanlong Zeng and Wensheng Gan are with the School of Intelligent Systems Science and Engineering, Jinan University, Zhuhai 519070, China. (E-mail: flzeng1@stu.jnu.edu.cn, wsgan001@gmail.com)}
    
\thanks{Kangjie Chen is with the School of Cybersecurity, Tianjin University, Tianjin 300354, China. (E-mail: kangjie\_chen@tju.edu.cn)}
    
\thanks{Philip S. Yu is with the University of Illinois Chicago, Chicago, USA. (E-mail: psyu@uic.edu)}

}

\maketitle

\begin{abstract}
    The class imbalance problem refers to the disproportionate distribution of samples across different classes within a dataset, where the minority classes are significantly underrepresented. This issue is also prevalent in graph-structured data. Most graph neural networks (GNNs) implicitly assume a balanced class distribution and therefore often fail to account for the challenges introduced by class imbalance, which can lead to biased learning and degraded performance on minority classes. We identify a quality inconsistency problem among synthetic nodes, leading to suboptimal performance under graph imbalance conditions. To mitigate this issue, we propose GraphIFE (Graph \underline{I}nvariant \underline{F}eature \underline{E}xtraction), a novel framework for improving the consistency of synthetic-node quality. Our approach incorporates two key concepts from graph invariant learning and introduces strategies to strengthen representations in the embedding space, thereby enhancing the model’s ability to identify invariant features. Extensive experiments on six benchmark datasets demonstrate the effectiveness and robustness of GraphIFE across different GNN backbones. The code is publicly available at \href{https://github.com/flzeng1/GraphIFE}{https://github.com/flzeng1/GraphIFE}.
\end{abstract}


\section{Introduction}\label{sec:Introduction}

Node classification is a fundamental task in the graph domain \cite{cook2006mining, ju2024comprehensive}. Graphs are widely used to model complex relational structures among entities, where nodes typically represent individual entities (such as individuals, objects, organizations, or events) and edges denote the relationships between them. For example, a social network can be modeled as a graph in which each node corresponds to a user, and the edges capture interactions or relationships among users. Node classification has a wide range of practical applications \cite{wu2022graph, yang2021consisrec}. In social networks, it can be used to detect malicious users \cite{ying2018graph, yang2025global, wei2025prompt}, while in protein–protein interaction networks, it facilitates the prediction of protein functions or annotations \cite{jumper2021highly}. Graph neural networks (GNNs) have emerged as a powerful tool to address node classification tasks and have shown strong performance in diverse graph-related applications \cite{wang2022powerful, wu2019simplifying}. However, most existing GNN approaches assume class-balanced distributions and therefore fail to adequately address the critical challenge of class imbalance in real-world graph-structured data \cite{zhao2021graphsmote, liu2023imbens, liu2025survey}. 

Class imbalance occurs when the quantity of samples varies significantly between classes, leading to biased learning and degraded performance, particularly for the minority class \cite{liu2025survey}. In real-world applications, graph-structured data frequently demonstrates significant class imbalance, where minority classes are substantially underrepresented relative to majority classes \cite{shchur2018pitfalls, galke2023lifelong}. A representative example emerges in Netflix's user-item interaction graph, where the node distribution exhibits pronounced structural imbalance: the cardinality of user nodes significantly exceeds video nodes by orders of magnitude. When applied to class-imbalanced graphs, conventional GNNs exhibit a systematic tendency to underrepresent minority classes, leading to biased representations and degraded generalization performance on underrepresented nodes \cite{chawla2002smote}. This limitation underscores the critical need for specialized techniques that explicitly address class imbalance in graph-based learning \cite{liu2025survey}, ensuring robust and equitable model performance across all classes. Recent, research on graph imbalance has produced numerous, which can be broadly categorized into two distinct types, which are node synthesis methods \cite{li2023graphsha, park2021graphens, liuclass} and loss-modified methods \cite{chen2021topology, liu2021tail, song2022tam}. Node synthesis methods aim to address class imbalance by augmenting the training data through the generation of synthetic nodes for minority classes. In contrast, loss-modification strategies address class imbalance by adaptively reweighting the loss function to prioritize minority classes during training, thereby correcting the training bias inherent in imbalanced datasets while preserving the original data distribution. In this paper, node synthesis is the focus of our research. After reviewing the entire process of node synthesis, we have found that the synthetic nodes exhibit the quality inconsistency problem. The problem means that the features of the synthetic nodes may raise a potential Out-Of-Distribution (OOD) issue. Based on the results of our experiment, we discovered that the problem would lead to performance degradation of the model for minority classification.

The primary challenge in resolving the quality inconsistency problem lies in mitigating the feature OOD of synthetic nodes. Although the objective is clear, resolving this issue remains formidable. Current methodologies lack an effective mechanism for ensuring that the synthetic nodes maintain fidelity to the original data distribution. Moreover, existing approaches struggle to precisely detect which synthetic nodes exhibit distributional divergence. The quality inconsistency problem in synthetic nodes manifests itself primarily in the following. The features of synthetic nodes exhibit a statistically distributed bias relative to the original training set. This bias propagates through the learning process as synthetic nodes dominate the minority class training, ultimately resulting in systematically distorted feature representations and degraded generalization performance for minority-class prediction. These challenges necessitate the development of robust methodological frameworks to effectively address them.

To address the challenge, we propose GraphIFE (Graph \underline{I}nvariant \underline{F}eature \underline{E}xtraction), a framework designed to mitigate quality inconsistencies among synthetic nodes. GraphIFE strategically circumvents the need to directly assess whether each synthesized node aligns with the original data distribution. It integrates concepts from invariant learning to directly extract the invariant node features while employing an adversarial attacker to enhance the model’s ability to distinguish among minority classes. We adopt two key concepts from graph invariant learning \cite{sui2024unleashing}: 1) Invariant features, which remain stable across different domains or structures and preserve predictive semantics under varying conditions. 2) Environment features, which vary across domains or graph structures but are not causally related to the prediction target. GraphIFE leverages both types of features to address these challenges. It introduces a representation alignment strategy that encourages semantically and topologically similar nodes to remain close in the embedding space. Additionally, it incorporates a gated mixing mechanism to support an aggressive learning strategy while simultaneously preventing noise from dominating the training process. We evaluate the effectiveness of our proposed method across a diverse range of benchmark datasets, including citation networks (Cora, Citeseer, and PubMed) \cite{sen2008collective} and Amazon co-purchase networks (Amazon-Computers, Amazon-Photo, and Coauthor-CS) \cite{shchur2018pitfalls}. These datasets are adapted to long-tailed and imbalanced settings, following GraphENS \cite{park2021graphens} and GraphSHA \cite{li2023graphsha} to better simulate real-world scenarios. Furthermore, we assess the generalizability of our approach across multiple GNN backbones, including GraphSAGE \cite{hamilton2017inductive}, GCN \cite{kipf2016semi}, and GAT \cite{velivckovic2017graph}. We also conduct experiments under different hyperparameter configurations to analyze their impact on model performance. In summary, the key contributions are as follows:

\begin{itemize}
    \item We identified the quality inconsistency issue in the node synthesis methods, where the features of synthetic nodes may exhibit potential OOD behavior, leading to suboptimal performance.
    
    \item We revisit the graph imbalance problem through the lens of invariant learning and propose GraphIFE, an adversarial feature perturbation framework. GraphIFE extracts invariant node representations and employs an environment feature extractor to construct challenging perturbations, thereby mitigating inconsistencies between synthetic and original nodes and improving minority-class discrimination.
    
    \item Through extensive experiments, we demonstrate that GraphIFE not only effectively mitigates the quality inconsistency issue but also achieves excellent performance across various datasets. It achieves competitive results across diverse datasets and GNN backbones.
\end{itemize}

Following the introduction, a brief review of related work on the graph imbalance problem and graph invariant learning is provided in Section \ref{sec: related work}. The details of the quality inconsistency issue are analyzed in Section \ref{sec: Quality Inconsistency}. Then, the preliminaries of this paper are provided in Section \ref{sec: preliminaries}, including the basic notation, the concise introduction of the node classification task, and the problem definition. The detailed implementation of GraphIFE is introduced in Section \ref{sec: methodologies}. The architecture and proposed mechanism are also introduced in this section. In Section \ref{sec: experiment}, we present an experimental evaluation to assess the effectiveness of GraphIFE. In addition to performance benchmarking, we conduct an in-depth analysis that includes ablation studies, visualizations, and an investigation of the sensitivity of GraphIFE to key hyperparameters. We also evaluate the performance of GraphIFE in large-scale graphs. Finally, we present our conclusions and outline future research work in Section \ref{sec: Conclusion}.

\section{Related Work} \label{sec: related work}

In this section, we provide a concise review of relevant prior work, including graph imbalance methods and graph invariant learning methods.

\subsection{Graph Imbalance Problem}

The class imbalance problem represents a pervasive challenge in real-world datasets, particularly in domains requiring complex pattern recognition, such as protein-molecule interaction studies \cite{japkowicz2002class, johnson2019survey}. This issue arises when the quantity of different classes exhibits a significantly uneven distribution, often leading models to bias predictions toward the majority class while underrepresenting the minority class \cite{park2021graphens, song2022tam}. Such an imbalance can substantially degrade model performance, particularly in classification tasks \cite{song2022tam}. While the class imbalance problem in graph-structured data shares similarities with its counterpart in conventional data settings, it also introduces unique complexities. In addition to disparities in class quantity, the graph class imbalance problem is characterized by topological imbalances, i.e., differences in structural connectivity between classes \cite{park2021graphens, song2022tam}. In graph-based learning, two major approaches have been developed to address this issue: node synthesis and loss modification \cite{chen2021topology, liu2021tail, song2022tam}. Node synthesis methods aim to mitigate class imbalance by generating synthetic nodes for minority classes, thus addressing the uneven quantity issue. For example, GraphSMOTE \cite{zhao2021graphsmote} synthesizes new minority nodes by interpolating features of two existing nodes from the same class, followed by an edge prediction step to preserve graph connectivity. GraphENS \cite{park2021graphens} constructs synthetic nodes by merging the ego-network of a minority node with that of a randomly selected node. GraphSHA \cite{li2023graphsha} introduces a more discriminative approach by generating nodes from hard-to-classify samples. Methods such as ImGAGN \cite{qu2021imgagn} and DR-GCN \cite{shi2020multi} apply Generative Adversarial Networks (GANs) \cite{goodfellow2020generative} to create synthetic nodes. However, many of these methods inadequately capture the intricate topological structures inherent in real graphs. In particular, synthesizing nodes from topologically ambiguous regions can hinder model performance, as the message-passing mechanisms in GNNs tend to propagate and amplify such noise. In contrast, loss modification methods adjust the training objective to compensate for class or topological imbalances. For instance, TAM \cite{song2022tam} introduces class-specific weights based on the distribution of node labels, while ReNode \cite{chen2021topology} adaptively modulates the contribution of each node by considering its proximity to class boundaries in the graph topology. GraphMuGu \cite{zhao2025generate} jointly synthesizes minority nodes and reweights authentic and generated samples to mitigate class imbalance in graph node classification. Additionally, other recent approaches leverage contrastive learning frameworks \cite{cui2023hybrid, zeng2023imgcl} or mathematical modeling techniques \cite{yan2024rethinking, zhu2023balanced} to tackle the graph class imbalance problem from alternative perspectives.

However, the above methods ignore the quality inconsistency of synthetic nodes during model training. Inconsistent quality of the synthesized nodes would result in suboptimal model performance. GraphIFE leverages the concept of invariant learning to mitigate the adverse effects of the quality inconsistency issue.

\subsection{Graph Invariant Learning}

Invariant learning methods aim to uncover stable and domain-invariant relationships between input features and target variables. Invariant learning aims to solve the problem of out-of-distribution (OOD). The OOD problem arises when the testing data distribution differs from, and is unknown relative to, the training distribution. Various strategies have been proposed to address OOD generalization \cite{liu2021towards, mo2024graph, jia2024graph}. Among them, causal inference methods seek to learn underlying variables within a causal graph in either an unsupervised or semi-supervised manner \cite{liu2023flood, wang2024dissecting}. By leveraging these learned causal representations, models can better capture the latent data-generating mechanisms. Graph OOD generalization poses greater complexity compared to its counterparts in other domains, as distributional shifts in graph data can manifest in diverse forms, including changes in node attributes, edge structures, or overall topology. These variations significantly complicate the identification and extraction of invariant features. The quality inconsistency of the mixed nodes found in this paper is also a graph OOD problem. The inconsistent quality of synthetic nodes is mainly manifested in the inconsistent features of synthetic nodes compared to the original dataset.

Currently, two primary categories of graph-related tasks are node-level and graph-level. They are typically addressed using distinct approaches. Node-level tasks focus on predicting labels for individual nodes, which are inherently non-independent and identically distributed (non-i.i.d.) due to the interconnected nature of graph structures. In contrast, graph-level tasks consider each entire graph as a single instance and often assume that these graph instances are i.i.d., despite potential underlying correlations across graphs. This article mainly discusses the node-level graph OOD problem. Many methods to deal with node-level OOD have been proposed.  SRGNN \cite{zhu2021shift} employs multiple context generators to construct diverse virtual environments and maximizes the variance of empirical risks across them. This encourages the model to explore a broader range of feature-context relationships, thereby enhancing its ability to extrapolate beyond the single observed training environment. EERM \cite{wu2022handling} is designed to address distributional discrepancies between biased training data and the true underlying inference distribution of the graph. FLOOD \cite{liu2023flood} integrates invariant representation learning with bootstrapped representation learning to enhance OOD generalization. The proposed approach seeks to strike a balance between maintaining stable representations across diverse training environments and preserving adaptability to potentially unknown test distributions. CIA-LRA \cite{wang2024dissecting} utilizes the distribution of neighboring labels to selectively align node representations. This approach enables the model to effectively identify and retain invariant features while filtering out spurious correlations, without requiring explicit environment labels.

However, existing methods remain limited in addressing graph imbalance scenarios. GraphIFE especially considers the graph imbalance node classification problem. It also considers topology imbalance, selects appropriate neighbors for the synthesized node, and extracts the essential features of the original nodes and synthetic nodes for training to mitigate the problem of graph imbalance.

\begin{figure*}
    \centering
        \subfloat[Minority Performance Gap]{
        \label{fig: motivation pre}
        \includegraphics[scale = 0.24]{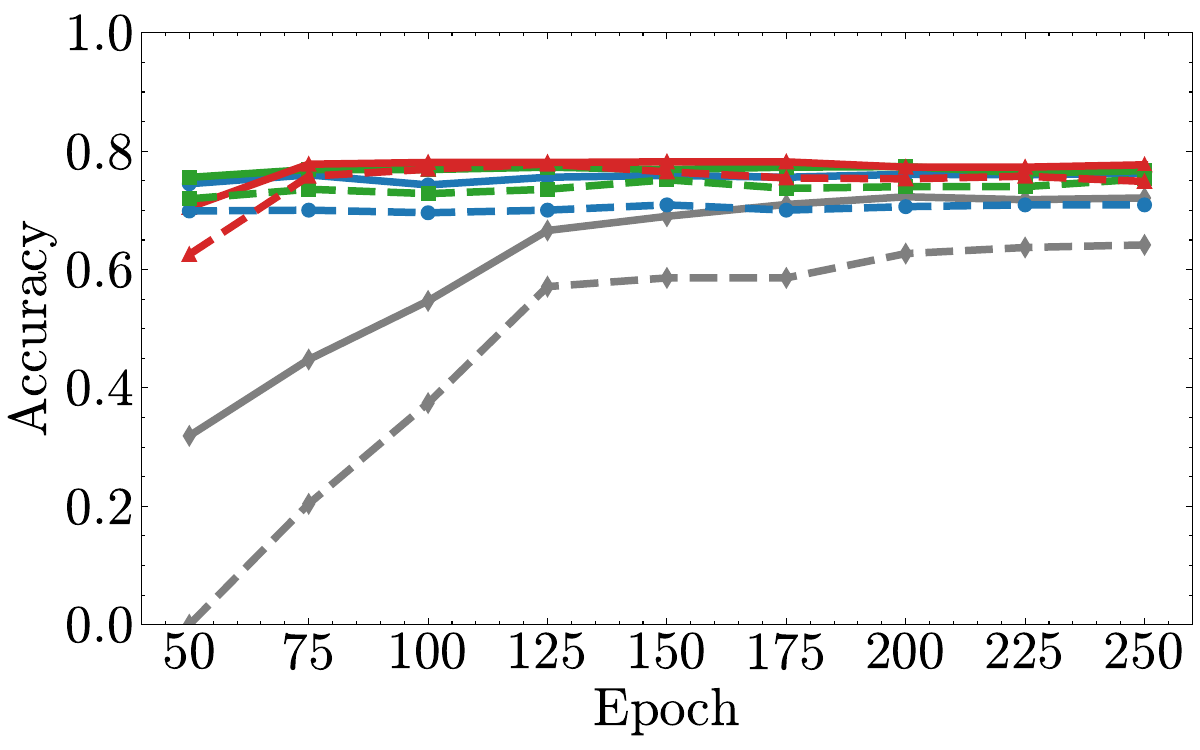}
    }
    \subfloat[Synthetic-Original Embedding MMD]{
        \label{fig: obs1 MMD in the Embedding}
        \includegraphics[scale = 0.25]{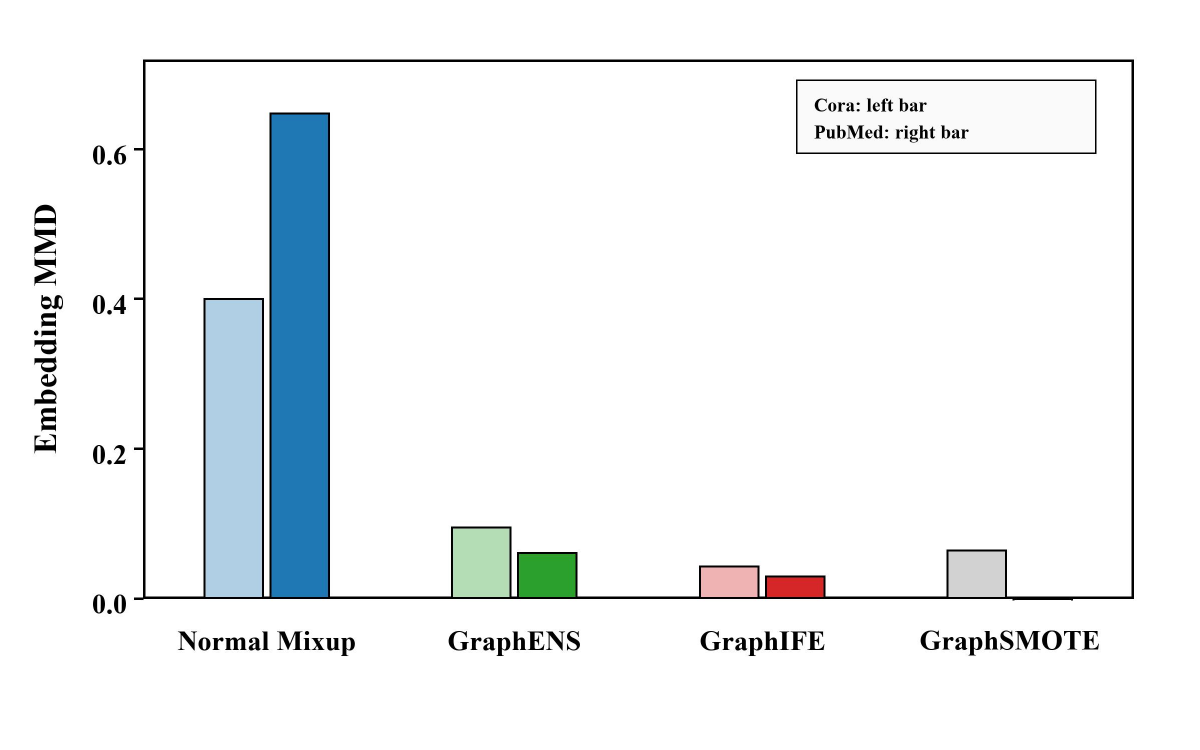}
    }
    \subfloat[Late-stage Minority Loss Share]{
        \label{fig: obs2 late_synthetic_loss_share}
        \includegraphics[scale = 0.25]{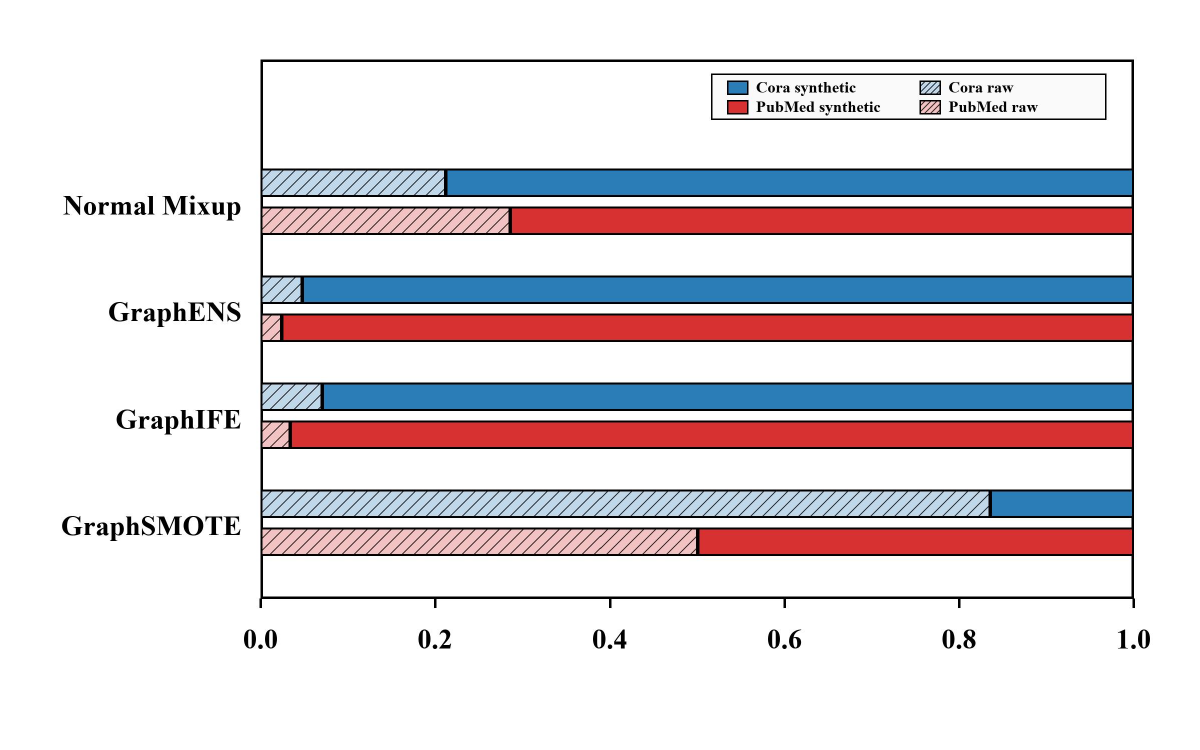}
    }
    \caption{Motivation analysis of synthetic-node quality inconsistency. 
(a) Test accuracy curves of all classes and minority classes. Dashed lines denote minority-class accuracy. 
(b) Embedding-space MMD between original minority nodes and synthesized minority nodes. Larger values indicate a stronger synthetic-original distribution discrepancy. 
(c) Late-stage loss contribution within the minority training objective. Dark bars denote synthetic nodes, and light bars denote original minority nodes.}
    \label{fig: motivation}
\end{figure*}

\section{Quality Inconsistency Problem} \label{sec: Quality Inconsistency}

In this section, we formally characterize the quality inconsistency problem in graph imbalance scenarios and provide an analysis of it.

\subsection {Rethinking Node Mixup} \label{sec: Rethinking Node Mixup}

To address the node classification problem in graphs with class imbalance, a crucial step is to balance the quantity of nodes across different classes. One classic method for balancing node quantities is node mixup, which synthesizes new nodes by combining existing nodes from all classes, thereby achieving a more balanced distribution across all classes. In recent years, a variety of node mixup methods have been proposed, including GraphENS \cite{park2021graphens}, GraphSMOTE \cite{zhao2021graphsmote}, and BAT \cite{yan2024rethinking}, among others. These methods employ diverse techniques to synthesize new nodes for the minority class, aiming to balance the graph and improve classification performance in imbalanced scenarios. Nevertheless, although these methods employ metrics to evaluate the rationality of synthetic nodes (such as using  KL divergence to measure the distance between synthetic nodes and original minority nodes), the quality of the synthesized minority nodes produced by these methods remains inconsistent.

This raises the question of how to mitigate the effects of quality inconsistency. This issue primarily arises from the distributional divergence between synthesized and original node features. To gain a deeper understanding of this phenomenon, we pose the following four questions: 
(1) \textbf{How does quality inconsistency affect empirical performance under class imbalance?}
(2) \textbf{How pronounced is the quality inconsistency between synthesized and original minority nodes?}
(3) \textbf{What mechanisms cause synthesized nodes to exert a
disproportionate influence on minority-class optimization?}
(4) \textbf{To what extent does GraphIFE mitigate this discrepancy while
preserving downstream discriminative performance?}

\subsection{Understanding Quality Inconsistency}
In this section, we present a detailed empirical study of the quality inconsistency problem, with the goal of understanding its underlying causes.

\textbf{Exploration}. Specifically, we perform a series of controlled experiments to systematically answer the four research questions formulated above. In our experiment, we use GCN as the GNN backbone, set the imbalance ratio to 100, and train each model for 250 epochs on the Cora and PubMed datasets to obtain the results. We follow the long-tailed setting proposed by \cite{park2021graphens} in our experiments.

\textbf{How does quality inconsistency affect empirical performance under class imbalance?} We analyze the test accuracy curves throughout the test process. As shown in Fig.~\ref{fig: motivation pre}, the solid lines denote the classification accuracy over all classes, whereas the dashed lines correspond to the classification accuracy of minority classes. The normal Mixup method generates synthetic nodes by linearly interpolating node features with a predefined mixing strategy.

A noticeable performance gap can be observed between the minority-class accuracy and the overall classification accuracy for several oversampling methods. In particular, GraphSMOTE exhibits a substantial discrepancy between these two curves. Similarly, GraphENS, Normal Mixup, and GraphIFE also show varying degrees of performance disparity, suggesting that merely balancing the number of training nodes does not necessarily lead to balanced generalization across classes. Among these methods, GraphIFE exhibits the smallest gap between minority-class and overall performance, indicating better generalization capability to minority nodes. Overall, these results suggest that the quality of the synthesized minority nodes is not sufficient to enable the model to classify the minority class effectively and motivate further investigation into whether synthetic minority nodes introduce distributional bias during training.

\textbf{How pronounced is the quality inconsistency between synthesized and original minority nodes?} We quantify the feature discrepancy between synthesized and original minority-class nodes using the Maximum Mean Discrepancy (MMD) computed on node representations in the embedding space. Fig.~\ref{fig: obs1 MMD in the Embedding} reports the MMD values on Cora and PubMed, where the darker and lighter bars correspond to Cora and PubMed, respectively. The results reveal a clear distributional discrepancy between synthesized and original node features across different node mixup methods. Since a larger MMD indicates a greater divergence between feature distributions, lower MMD values imply better feature consistency. Among all methods, Normal Mixup exhibits substantially larger MMD values, suggesting that the generated node features deviate considerably from the original minority-node distribution. In contrast, others generally produce more consistent synthetic features. In particular, GraphIFE achieves the lowest MMD on Cora, while GraphSMOTE obtains the lowest MMD on PubMed, indicating that their synthesized features are more closely aligned with those of the original minority nodes in the embedding space. Note that MMD only indicates distributional proximity and does not necessarily imply discriminative utility. The results show that synthetic nodes generated by existing mixup methods are not uniformly aligned with original minority nodes, supporting the existence of synthetic-original quality inconsistency.

\textbf{What mechanisms cause synthesized nodes to exert a disproportionate influence on minority-class optimization?} We further investigate the underlying cause of quality by analyzing the proportion of training loss contributed by synthetic and original minority nodes during the late stage of training. As shown in Fig.~\ref{fig: obs2 late_synthetic_loss_share}, the darker bars correspond to synthetic nodes, while the lighter bars represent original minority nodes. Except for GraphSMOTE, the optimization process is predominantly driven by synthetic nodes, whose loss contribution substantially exceeds that of the original nodes.

\begin{figure}
    \centering
    \subfloat[Majority-Minority F1 Gap]{
        \label{fig: majority_minority_f1_gap}
        \includegraphics[scale = 0.24]{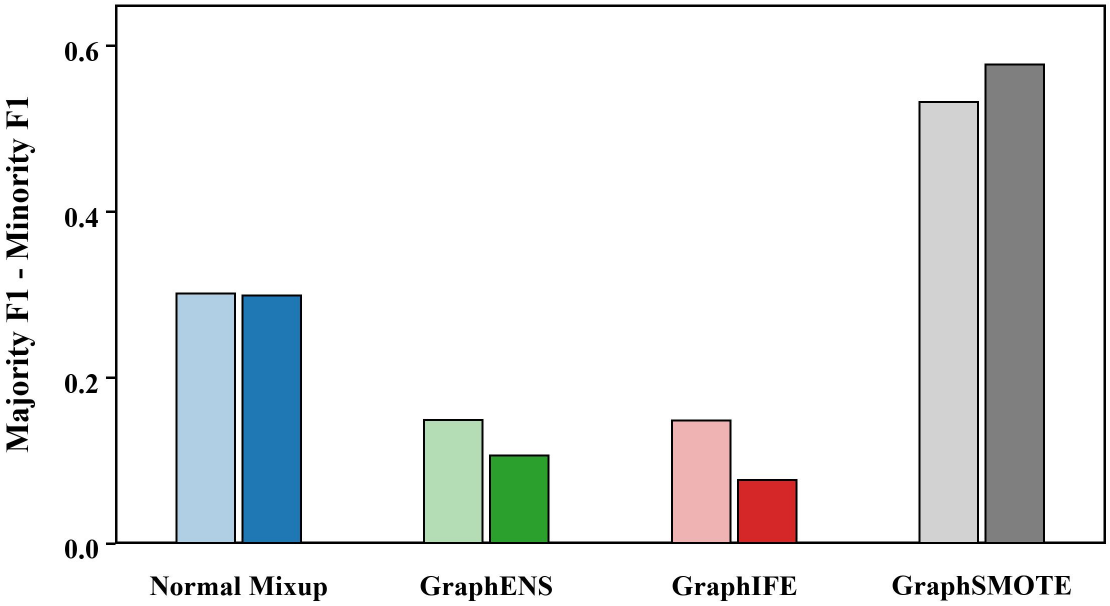}
    }
    \subfloat[Final Accuracy]{
        \label{fig: motivation final acc}
        \includegraphics[scale = 0.24]{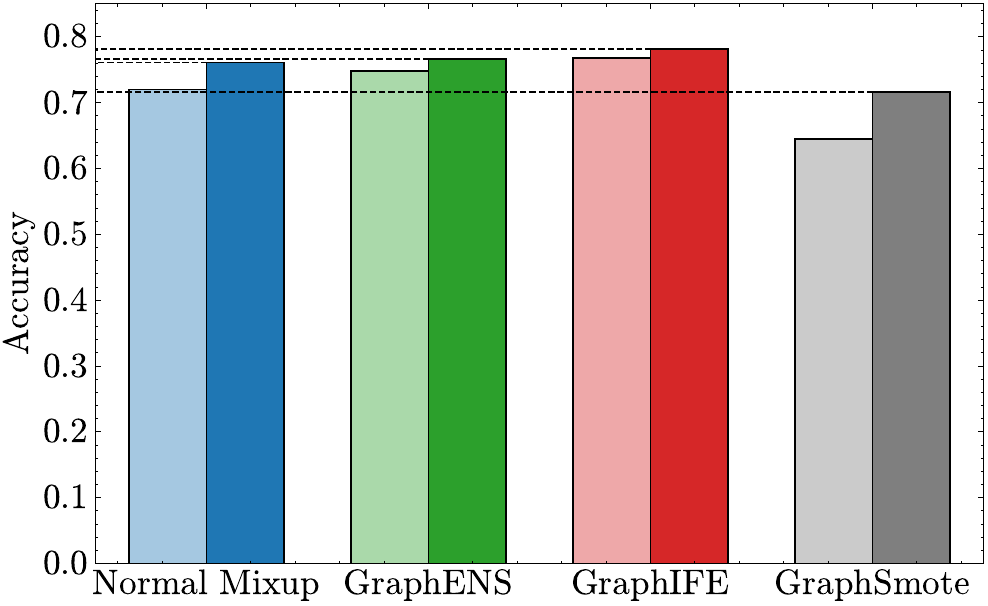}
    }

    \caption{Performance consequences of synthetic-node quality inconsistency.
    (a) reports the majority-minority F1-score gap on Cora and PubMed, where a smaller gap indicates more balanced performance between majority and minority classes.
    (b) reports the final classification accuracy of different node mixup methods.
    Dark bars denote all-class test accuracy, while light bars denote minority-class test accuracy. The final checkpoint is selected according to validation accuracy.
    }
    
    \label{fig: motivation_result}
\end{figure}

\textbf{To what extent does GraphIFE mitigate this discrepancy while preserving downstream discriminative performance?} As illustrated in Fig. \ref{fig: majority_minority_f1_gap}, all compared methods exhibit varying degrees of F1-score disparity between majority and minority classes.  Among them, GraphIFE achieves the smallest F1-score gap, indicating the most balanced performance across majority and minority classes.

\begin{figure}
    \centering
    \includegraphics[width=0.34\textwidth]{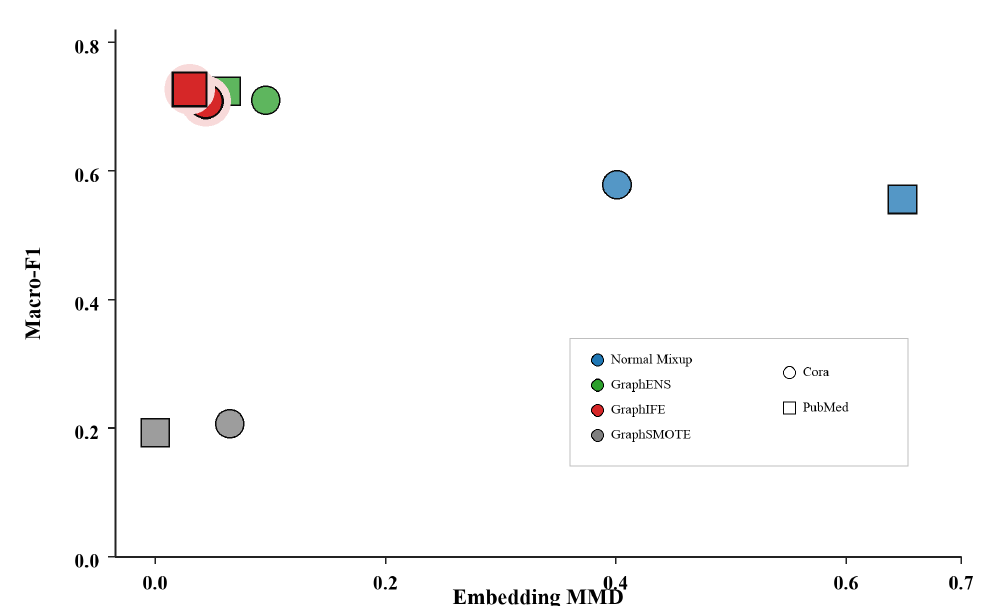}
    \caption{Alignment vs. utility. The x-axis reports the embedding-space MMD between original and synthesized minority nodes, and the y-axis reports Macro-F1. GraphSMOTE achieves a small MMD but low Macro-F1, indicating that synthetic-original distributional proximity alone is insufficient for effective minority classification.    }
    \label{fig: alignment_utility}
\end{figure}

As shown in Fig.~\ref{fig: motivation final acc}, the darker bars represent the test accuracy over all classes, whereas the lighter bars correspond to the test accuracy of minority classes. It can be observed that Normal Mixup and GraphSMOTE exhibit a considerable performance disparity between minority-class and overall classification accuracy. In contrast, GraphIFE achieves a much smaller discrepancy, indicating a better alignment between minority-class learning and overall model performance. 
As a complementary observation, GraphSMOTE provides an informative counterexample. 
It is worth noting that MMD only measures the distributional proximity between original and synthesized minority nodes, but does not directly characterize the discriminative utility of the learned representations. As shown in Fig. \ref{fig: alignment_utility}, GraphSMOTE obtains a small embedding MMD but still suffers from low Macro-F1. This indicates that reducing synthetic-original discrepancy alone is insufficient; effective augmentation should also preserve discriminative information that benefits downstream minority classification.

In summary, the observations above reveal a coherent failure mechanism for node mixup under class imbalance. Although node mixup balances the number of minority nodes, synthetic minority nodes may remain distributionally inconsistent with original minority nodes. Since synthetic nodes can dominate the late-stage minority optimization objective, such inconsistency may become a major training signal and lead to degraded minority-class generalization. GraphIFE is therefore designed to learn invariant and discriminative representations that reduce the harmful effect of synthetic-original inconsistency while preserving downstream utility.

\subsection{Analysis}

The quality inconsistency problem primarily arises from distributional inconsistencies in synthetic-node features. These inconsistencies distort node representations, confuse the GNN during training and degrade its performance. The issue is especially pronounced for the minority classes, causing the GNN to perform poorly in imbalanced node classification. A theoretical analysis of these aspects is provided in the following section.

Initially, we define the notation in this analysis. Given a training dataset, $\mathcal{D}_{tr}$ = \{$X_{c_1}$, ..., $X_{c_n}\}$, $X_{ci}$ denotes the feature set of the $c_i$ class. In the imbalanced situation, the training set can be simply divided into $\mathcal{D}_{tr}$ = $\{X_{min}, X_{maj} \}$, which means the training set is composed of the feature set of the minority class and the feature set of the majority class. The label set can be divided into $\mathcal{Y}$ = \{$c_1$,.., $c_n$\} = $\{c_{min}$, $c_{maj}\}$ in the same way. Additionally, the letter with an overline above denotes the feature of a synthesized node, such as $\bar{x}$ denoting the feature of a synthesized node. The quality inconsistency issue especially affects the minority class. This is due to the quantity of the synthesized minority nodes being far more than the original minority nodes. To analyze the issue, we have the following definition:

\begin{definition} [Quality Inconsistency]
\rm The issue of synthetic nodes would confuse the GNN model, making the performance of the model unable to distinguish the minority class. Its mathematical expression is as follows:
    \begin{equation}
        {\rm FI}(P_{X_{syn}}, P_{X_{ori}}) = \frac{1}{2} \int_S |P_{X_{syn}}(k) -  P_{X_{ori}}(k)|dk,
    \end{equation}

    where $S$ = $\{k \in X |\mathcal{Y}_k \in c_{min}\}$, which is the feature of the minority class after node mixup. $P$ is a distribution. $FI(\cdot, \cdot)$ is always mathematically bounded within [0, 1].
\end{definition}

The formulation above indicates that for the minority class, the feature distribution of the synthetic nodes is potentially different from the original training set. The definition characterizes synthetic-original inconsistency using total variation distance (TVD) for conceptual clarity. In the empirical analysis, we use maximum mean discrepancy (MMD) as a kernel-based proxy for estimating distributional difference from finite samples, since TVD is difficult to estimate reliably in high-dimensional representation spaces.

The features of the synthesized node have a significant impact on the representation of the minority class and can dominate the training process. The goal of imbalanced node classification is to balance an imbalanced dataset. Initially, quantitative parity is ensured to be equal by node mixup. To the majority class, it has $|X_{maj}| \gg |\bar{X}_{maj}|$. On the contrary, for the minority class, it has $|X_{min}| \ll |\bar{X}_{min}|$. The letter with an overline above means it belongs to the synthesized node. In the node mixup process, there is always a mixup method to synthesize nodes. To a synthesized node $\bar{v}$, the feature of $\bar{v}$ is synthesized as:
\begin{equation}
\begin{split}
    \bar{x} & = \text{MIXUP($x_{sample}$, $x_{target}$)} \\
    & = px_{sample} + (1-p)x_{target}.
\end{split}
\end{equation}
$p$ is the mix ratio of the different node mixup methods. The above equation means that, in the node mixup method, there always exists a different ratio to determine the mix proportion of the candidate nodes. The question that exists in this process is whether the synthesized feature would potentially be out of distribution and lack stable quality. Additionally, due to $|X_{min}| \ll |\bar{X}_{min}|$, the whole training process of the minority class would be dominated by the synthesized nodes. We quantify the dominance of synthetic nodes in minority optimization by the synthetic loss share:
\begin{equation}
    \rho_{\mathrm{syn}}
=
\frac{
\sum_{v_i \in \mathcal{V}^{\mathrm{syn}}_{\min}} \ell_i
}{
\sum_{v_i \in \mathcal{V}^{\mathrm{syn}}_{\min}} \ell_i
+
\sum_{v_i \in \mathcal{V}^{\mathrm{ori}}_{\min}} \ell_i
}.
\end{equation}
A larger \(\rho_{\mathrm{syn}}\) indicates that the minority-class training objective is mainly shaped by synthetic minority nodes. When \(\rho_{\mathrm{syn}}\) is large, the optimization signal of the minority class is mainly determined by synthetic minority nodes.

Without stable feature quality, the model's performance for the minority class would be affected. The feature quality inconsistency problem can be formally represented through the following mathematical process:

The resulting balanced dataset satisfies the following property:
$|X'_{\mathrm{maj}}| = |X'_{\mathrm{min}}|
= |X_{\mathrm{maj}} \cup \bar{X}_{\mathrm{maj}}|
= |X_{\mathrm{min}} \cup \bar{X}_{\mathrm{min}}|.$

For the majority class, since
$|X_{\mathrm{maj}}| \gg |\bar{X}_{\mathrm{maj}}|,$
training is primarily dominated by the original node features. In contrast, for the minority class, since
$|X_{\mathrm{min}}| \ll |\bar{X}_{\mathrm{min}}|$,
training is largely dominated by synthesized node features. Moreover, each synthesized feature is generated as
$\bar{x} = p x_{\mathrm{sample}} + (1-p)x_{\mathrm{target}}$,

which generally implies \(\bar{x} \neq x\). Therefore, for the GNN model, the empirical risk over the balanced training distribution can be written as
\begin{equation}
    \hat{f}_{\theta}
=
\arg\min_{\theta}
\mathbb{E}_{(v,c_{\mathrm{min}})\sim \mathcal{D}'_{\mathrm{tr}}}
\left[
\mathcal{L}\left(\hat{f}_{\theta}(v), c_{\mathrm{min}}\right)
\right].
\end{equation}
Furthermore, let
\[
a = \frac{|X_{\mathrm{ori}}|}{|X_{\mathrm{syn}}| + |X_{\mathrm{ori}}|},
\]
where \(\hat{f}_{\theta}(v)=\mathrm{Proj}(h_v^{(l)})\), and \(h_v^{(l)}\) denotes the representation of node \(v\) at layer \(l\).

Ignore the number of layers, $\phi$ denotes $\texttt{AGG}$, to the representation of the minority node $v$ in the balanced dataset:

\begin{equation}
\begin{split}
   h_v & = a\phi[\frac{1}{|u|}\sum_{u\in \mathcal{N}(v)}h_u, h_{v}] + (1-a)\phi[\frac{1}{|t|}\sum_{t\in \mathcal{N}(\bar{v})}h_u, h_{\bar{v}}] \\
   & \approx \phi[\frac{1}{|t|}\sum_{t\in \mathcal{N}(\bar{v})}h_u, h_{\bar{v}}], \,\, (a = \frac{|X_{ori}|}{|X_{syn}| + |X_{ori}|} \to 0)
\end{split}
\end{equation}

The whole representation of the minority class is dominated by the synthesized minority node. To the model, the above issue results in 
\begin{equation}
\small
\begin{split}
       & \mathbb{E}_{(u, c_{min})\sim \mathcal{D}^{\prime}_{tr}}[\mathcal{L}(\hat{f}_\theta(u)), c_{min}] \\
    = &
    \mathbb{E}_{(v, c_{min})\sim \mathcal{D}^{\prime}_{tr}}[\mathcal{L}(\hat{f}_\theta(v)), c_{min}] + \mathbb{E}_{(\bar{v}, c_{min})\sim \mathcal{D}^{\prime}_{tr}}[\mathcal{L}(\hat{f}_\theta(\bar{v})), c_{min}]  \\
    \approx &  \mathbb{E}_{(\bar{v}, c_{min})\sim \mathcal{D}^{\prime}_{tr}}[\mathcal{L}(\hat{f}_\theta(\bar{v})), c_{min}]
\end{split}
\end{equation}

$\mathcal{D}^{\prime}_{tr}$ denotes the balanced training set after node mixup, and $u \in \mathcal{D}^{\prime}_{tr}$. Therefore, the classification ability of the model for the minority class tends to be completely trained by the synthesized node.

\section{Preliminaries} \label{sec: preliminaries}

In this section, we introduce the notation and core concepts used in this paper. The overview of GNN for node classification and problem definition is also introduced here.  

\subsection{Notation and Concept}

In this paper, we focus on the node classification task performed on an undirected graph 
$\mathcal{G}$ = $(\mathcal{V}, \mathcal{X}, \mathcal{A}, \mathcal{Y})$, where $\mathcal{V} = \{v_1, v_2, \ldots, v_n\}$ denotes the set of $n$ nodes. The node features are represented by the matrix 
$\mathcal{X} \in \mathbb{R}^{n \times d}$, 
where $d$ is the dimensionality of the feature space, and $\mathcal{X}_i$ (the $i$-th row of $\mathcal{X}$) corresponds to the feature vector of node $v_i$. The graph structure is encoded in the adjacency matrix 
$\mathcal{A} \in \mathbb{R}^{n \times n}$. 
The set of class labels is denoted by 
$\mathcal{Y}$ = \{1, 2, $\ldots$, $c$\}, 
comprising $c$ classes, where $\mathcal{Y}_i$ represents the $i$-th class, and $\mathcal{Y} (v_i)$ denotes the label assigned to node $v_i$. 
We also use $v_{i,j}$ to refer to node $v_i$ with the label $j$. Furthermore, we define $\mathcal{N} (v_i)$ as the set of one-hop neighbors of the node $v_i$. 
Additionally, we define the imbalance ratio $\rho$ as the ratio of the number of nodes in the most frequent class to the number of nodes in the least frequent class, given by  $\rho = \frac{|\{v \in \mathcal{V} | \mathcal{Y}(v) = \mathcal{Y}_{most freq}\}|}{|\{v \in \mathcal{V} | \mathcal{Y}(v) = \mathcal{Y}_{least freq}\}|}$. $\mathcal{Y}_{\textit{most freq}}$ denotes the class with the highest frequency of occurrence, that is, the class with the largest quantity. $\mathcal{Y}_{\textit{least freq}}$ denotes the class with the least frequency of occurrence. This ratio provides a measure of the class imbalance within the graph. Moreover, GraphIFE leverages both features to address the imbalance: 1) invariant features, which remain stable across different domains or structures and preserve predictive semantics under varying conditions; and 2) environment features, which vary across domains or graph structures but are not causally related to the prediction target. 

\subsection{GNN for Node Classification}

Node classification is one of the fundamental tasks in graph-based machine learning, where the objective is to predict the labels of nodes based on their features and the graph topology. Given a graph $\mathcal{G}$ = $(\mathcal{V}, \mathcal{X}, \mathcal{A}, \mathcal{Y})$, and a set of labeled nodes $\mathcal{V}_L \subset \mathcal{V}$, the goal is to infer the labels of the remaining unlabeled nodes $f (\mathcal{X}, \mathcal{A}) \rightarrow \hat{\mathcal{Y}} \in \mathbb{R}^{N \times C}$.

GNNs have emerged as powerful models for the node classification task by effectively combining node features with topological information through message-passing mechanisms. At each layer, each node aggregates information from its neighbors and updates its representation. For a given node $v$, its representation $h_v^{(l)}$ at the $l$-th layer of GNN is updated according to the following formulation:
\begin{equation}
    h_v^{(l)} = \texttt{UPDATE}\left(h_v^{(l-1)},\ \texttt{AGG}\left(\left\{ h_u^{(l-1)} \mid u \in \mathcal{N}(v) \right\}\right)\right),
\end{equation}
where $h_v^{(l)}$ denotes the representation of the node $v$ at the $l$-th layer, and $h_v^{(l-1)}$ is the representation from the previous layer. The function \texttt{AGG} performs an aggregation over the set of neighboring node representations $\left\{ h_u^{(l-1)} \mid u \in \mathcal{N}(v) \right\}$, where $\mathcal{N}(v)$ denotes the set of immediate neighbors of node $v$. The \texttt{UPDATE} function then combines the node's previous representation $h_v^{(l-1)}$ with the aggregated neighborhood information to produce the updated embedding $h_v^{(l)}$. It is noticed that the initial node representation is defined as $h_v^{(0)}$ = $X_v$, where $X_v$ is the input feature vector of the node $v$. 

After computing the final-layer node embeddings, a projector is applied to map each embedding to a prediction vector $\hat{y}_i$ over the target classes. The model is then trained using  cross-entropy \cite{liu2025survey} to compute over the labeled nodes in the training set $V_{\text{train}}$:
\begin{equation}
\mathcal{L} = \frac{1}{|V_{\text{train}}|} \sum_{i \in V_{\text{train}}} \text{CrossEntropy}(\hat{y}_i, y_i),
\end{equation}
where $y_i$ denotes the ground-truth label of node $i$, and $\hat{y}_i$ is the predicted class label output by the model.

\subsection{Problem Definition}

The primary objective of imbalance graph node classification is to address the class imbalance problem by synthesizing minority class nodes, thereby creating a more balanced graph structure and mitigating the detrimental effects of class imbalance on the original graph $\mathcal{G}$. To achieve this, a data augmentation strategy generates an augmented graph $\mathcal{G}'$. This augmented graph is then used as input for a GNN that performs a standard node classification task. By leveraging the augmented graph $\mathcal{G}'$, the GNN can learn more robust and balanced node representations. In this paper, GraphIFE extracts the invariant features of both synthesized and original nodes to mitigate the inconsistent quality of the synthetic nodes.

\begin{figure*}[htbp]
    \centering
    \vspace{-1.0cm} 
    \includegraphics[scale=0.55]{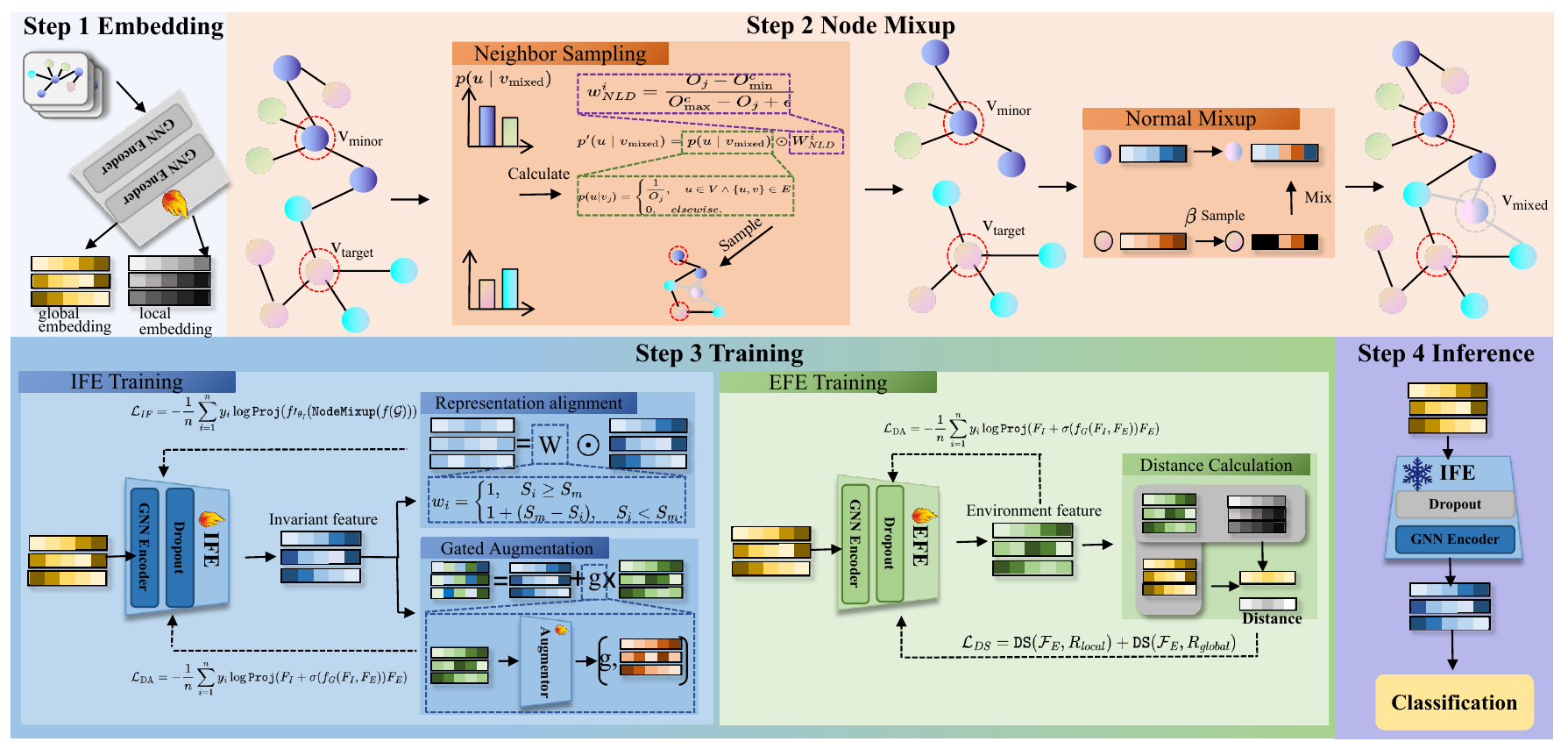} 
     \caption{The overview of the GraphIFE framework. GraphIFE is optimized in an alternating manner. The training process consists of two distinct phases: (1) the training of the invariant feature extractor and (2) the training of the environment feature extractor. 
     Initially, the input imbalanced graph is processed through a shared GNN encoder, which simultaneously generates both local node embeddings and global graph embeddings.
     In phase (1), the invariant feature $\mathcal{F}_I$ is extracted by \textit{IFE}; Then, the EFE extracts the environment features $\mathcal{F}_E$ of the graph. $\mathcal{F}_I$ and $\mathcal{F}_E$ are combined through the gated augmentor to produce mixed features $\mathcal{F}_m$.
     The final loss computation incorporates both $\mathcal{F}_m$ and $\mathcal{F}_I$, where the invariant feature loss undergoes representation alignment post-processing. 
     In phase (2), $\mathcal{F}_E$ of the global embedding is extracted by EFE. Then, the local and global embeddings both calculate the MSE distance with $\mathcal{F}_E$. The final loss function incorporates both $\mathcal{F}_E$ and MSE distance results.}
    \label{fig: GraphIFE}
\end{figure*}

\section{Methodologies} \label{sec: methodologies}

This section details the GraphIFE architecture, including its core components and the employed training strategies. The key idea of GraphIFE is not to generate more minority nodes, but to reduce the harmful effect of quality inconsistency by learning invariant and discriminative representations under synthetic-node-dominated optimization. It comprises two key components: an invariant feature extractor and an environmental feature extractor. The overall architecture of the framework is depicted in Fig. \ref{fig: GraphIFE}.

\subsection{Implementations of GraphIFE}

GraphIFE consists of the following components: (1) a two-layer GNN that encodes the input graph data into node-level embeddings; (2) an invariant feature extractor extracts invariant features from the embeddings; (3) an environment feature extractor extracts the environment features from the embeddings; (4) a gated augmentor, which employs a gating mechanism to adaptively combine the invariant and environmental features for data augmentation; and (5) a projector that maps the augmented features to the output space for downstream tasks. Note that the data augmentation strategy draws partial inspiration from AIA \cite{sui2024unleashing}. While GraphIFE performs feature-level augmentation by introducing controlled noise to node attributes, AIA operates at the graph-level for covariate shift.

We detail the architecture of the feature extractors and describe the complete process of feature extraction within the proposed framework. The invariant feature extractor and the environment feature extractor share a common architectural framework $M$, but have separate parameter sets, $\theta_I$ and $\theta_E$, respectively. Each extractor is optimized using a distinct objective function that align with its specialized roles in feature representation. The architecture of the feature extractor is based on a one-layer GNN denoted as  $f'$. Given an imbalanced input graph $\mathcal{G}$, a GNN encoder $f$ first processes the input to generate node representations $Z$. These representations are then utilized to synthesize new nodes and select appropriate neighbors, resulting in a transformed and more balanced graph $\mathcal{G}'$, from which balanced node representations $Z'$ are derived. Subsequently, the feature extractor $M$ is applied to both the original and synthetic nodes in $Z'$ to extract the final set of features $\mathcal{F}$.

The overall feature extraction process can be formally summarized as follows:
\begin{equation}
    Z = f(\mathcal{G}), \quad Z'= \text{Node Mixup}(Z), \quad \mathcal{F} = f'(Z').
\end{equation}

It is important to note that Node Mixup refers to the entire balancing procedure, which encompasses both node synthesis and neighbor selection.

\subsection{Invariant Feature Extractor}

The invariant feature extractor is designed to capture the stable, environment-invariant features of each node, thereby mitigating potential quality inconsistency introduced by synthetic nodes. Incorporating the invariant feature extractor enhances the model’s capability to effectively distinguish minority class nodes during training. The whole process can be summarized as: $\mathcal{F}_I $ = $f'_I(\text{Node Mixup}(f(\mathcal{G})))$ where $f(\mathcal{G})$ denotes the initial encoding of the input graph, and \(\text{Node Mixup}(\cdot)\) represents the node synthesis and neighbor selection process. Subsequently, the invariant feature representation $\mathcal{F}_I$ is passed through a projector to generate the predicted outputs. These predictions are then compared against the ground-truth labels using the loss function, yielding the per-node invariant loss vector defined as:
\begin{equation}
\ell_{IF,i}
=
-\mathbf{y}_i^{\top}
\log \left(
{\rm \texttt{Proj}}\left(
f'_{\theta_I}\left(
{\rm \texttt{NodeMixup}}\left(
f(\mathcal{G})
\right)
\right)_i
\right)
\right).
\label{eq:invariant_node_loss}
\end{equation}
Here $n$ denotes the number of samples, $y_i$ represents the ground truth, and \textit{Proj} denotes a projector that is applied to map each embedding to a prediction vector.

\begin{algorithm}[t]
\small
\caption{Representation alignment.}
\label{alg:representation_alignment}
\KwIn{Node representations $\mathbf{H}\in\mathbb{R}^{N\times d}$,
labels $\mathbf{y}$, and number of classes $C$}
\KwOut{Alignment weights $\mathbf{w}\in\mathbb{R}^{N}$}

Initialize $\mathbf{w}\leftarrow\mathbf{1}$\;

\For{$c\leftarrow 1$ \KwTo $C$}{
    $\mathcal{I}_{c}\leftarrow\{i\mid y_i=c\}$\;

    \If{$|\mathcal{I}_{c}|>1$}{
        $\boldsymbol{\mu}_{c}\leftarrow
        \frac{1}{|\mathcal{I}_{c}|}
        \sum_{i\in\mathcal{I}_{c}}\mathbf{h}_{i}$\;

        $s_i\leftarrow
        \frac{\mathbf{h}_{i}^{\top}\boldsymbol{\mu}_{c}}
        {\|\mathbf{h}_{i}\|_{2}\|\boldsymbol{\mu}_{c}\|_{2}}$,
        $\ \forall i\in\mathcal{I}_{c}$;\quad
        $\bar{s}_{c}\leftarrow
        \frac{1}{|\mathcal{I}_{c}|}
        \sum_{i\in\mathcal{I}_{c}}s_i$\;

        $w_i\leftarrow 1+\max(0,\bar{s}_{c}-s_i)$,
        $\ \forall i\in\mathcal{I}_{c}$\;
    }
}
\Return $\mathbf{w}$\;
\end{algorithm}

\textbf{Representation alignment.} For the minority class, another critical challenge lies in the quality disparities among synthesized node representations during training. This variation can hinder model stability and reduce the effectiveness of synthetic nodes in mitigating class imbalance. To address this issue, we propose a representation alignment mechanism. Specifically, GraphIFE computes an adaptive weight for the invariant feature loss associated with each node. This weighting strategy mitigates representation variance within the minority class, thereby enhancing the consistency and quality of the synthesized node representations. GraphIFE first computes the mean feature representation for each class. Subsequently, it calculates the similarity score $S_i$ between each intra-class node and its corresponding class mean. To mitigate the quality disparities, nodes with similarity scores below the class-wise average similarity $S_m$ are assigned a weight of 1 + $(S_m - S_i)$. This adaptive weighting scheme enables the model to focus more on intra-class nodes with greater representational variance. The weight $W$ assigned to each node is defined as follows:
\begin{equation}
w_i = \left\{
\begin{aligned}
& 1,  \quad S_i \ge S_m\\
& 1 + (S_m - S_i), \quad S_i < S_m.\\
\end{aligned}
\right.
\end{equation}

Finally, the representation-aligned node-wise invariant loss vector is
computed as \(\boldsymbol{\ell}'_{IF}
=\boldsymbol{\ell}_{IF}\odot\mathbf{W}\), where
\(\mathbf{W}=[w_1,w_2,\ldots,w_n]^{\top}\) and \(\odot\) denotes the
Hadamard product. The scalar representation-aligned invariant loss is then
computed as:
\begin{equation}
\mathcal{L}_{IF}
=
\frac{1}{n}
\sum_{i=1}^{n}
\ell'_{IF,i}.
\label{eq:aligned_invariant_loss}
\end{equation}

The pseudocode is described in Algorithm \ref{alg:representation_alignment}. 

\begin{algorithm}[t]
\small
\caption{NLD-guided neighbor weighting.}
\label{alg:nld_weight}
\KwIn{Graph $\mathcal{G}$, labels $\mathbf{y}$, training node set
$\mathcal{T}$}
\KwOut{NLD-guided weights
$\boldsymbol{\omega}^{\mathrm{NLD}}\in\mathbb{R}^{|\mathcal{V}|}$}

Initialize
$\boldsymbol{\omega}^{\mathrm{NLD}}\leftarrow\mathbf{0}$\;

\For{each training node $v_j\in\mathcal{T}$}{
    Compute its node-wise label homophily
    $O_j$ using its training neighbors\;
}

\For{each class $c=1,\ldots,C$}{
    $\mathcal{T}_c\leftarrow\{v_j\in\mathcal{T}\mid y_j=c\}$\;

    $O_{\min}^{c}\leftarrow\min_{v_j\in\mathcal{T}_c}O_j$\;
    $O_{\max}^{c}\leftarrow\max_{v_j\in\mathcal{T}_c}O_j$\;

    \For{each node $v_j\in\mathcal{T}_c$}{
        $\omega_j^{\mathrm{NLD}}
        \leftarrow
        1+
        \frac{O_j-O_{\min}^{c}}
        {O_{\max}^{c}-O_{\min}^{c}+\epsilon}$\;
    }
}

\Return $\boldsymbol{\omega}^{\mathrm{NLD}}$\;
\end{algorithm}

\textbf{Neighbor sampling.} Neighbor sampling is a strategy used to determine the set of nodes to which a newly synthesized node, denoted as $\bar{v}$, should be connected. The primary objective of this approach is to expose minority-class nodes to a more diverse set of structural and semantic neighborhoods. In this work, we construct the adjacent node distribution $p(u \!\!\mid\!\! \bar{v})$ following the approach used by in GraphENS \cite{park2021graphens}. In addition, we extend this formulation by incorporating the neighbor label distribution (NLD), which captures the label distribution within a node’s local neighborhood. While conventional approaches account only for first-order neighborhood frequencies, they fail to leverage the structural information encoded in the NLD. We argue that incorporating the NLD is crucial for constructing a more informative and balanced sampling distribution, as it can help address the neighbor memorization problem highlighted by GraphENS, wherein the model tends to overfit to the labels of frequently encountered neighbors. We formally define the NLD-Guided Node Weight as follows:

\begin{definition}[NLD-Guided Node Weight]
\rm For each training node $v_j \in \mathcal{T}$, we calculate NLD-Guided Node Weight as:
\begin{equation}
O_j =
\frac{1}{|\mathcal{N}_{\mathcal{T}}(v_j)|+\epsilon}
\sum_{v_k \in \mathcal{N}_{\mathcal{T}}(v_j)}
\mathbb{I}(y_j=y_k),
\label{eq:node_homophily}
\end{equation}
where $\mathcal{N}_{\mathcal{T}}(v_j)=\mathcal{N}(v_j)\cap\mathcal{T}$
denotes the training neighbors of $v_j$. The NLD-guided weight is
computed by class-wise Min-Max normalization in Eq.~\eqref{eq:nld_weight}.
\end{definition}

For each class \(c\), let
\(\mathcal{T}_c=\{v_j\in\mathcal{T}\mid y_j=c\}\),
\(O_{\min}^{c}=\min_{v_j\in\mathcal{T}_c}O_j\), and
\(O_{\max}^{c}=\max_{v_j\in\mathcal{T}_c}O_j\). The NLD-guided weight
of a training node \(v_j\in\mathcal{T}_c\) is computed as
\begin{equation}
\omega_j^{\mathrm{NLD}}
=
1+
\frac{O_j-O_{\min}^{c}}
{O_{\max}^{c}-O_{\min}^{c}+\epsilon}.
\label{eq:nld_weight}
\end{equation}
For nodes outside the training set, we set
\(\omega_j^{\mathrm{NLD}}=0\). The pseudocode is described in Algorithm \ref{alg:nld_weight}.

\subsection{Adversarial Environment Feature Extractor}

The environment feature extractor is used to obtain the environment features of each node. These extracted features are subsequently employed to introduce controlled perturbations that challenge the model, thereby enhancing its ability to distinguish class boundaries with greater precision, particularly for minority classes. Specifically, the environment feature extractor derives the embedding representation of environment features within the latent space, formally expressed as $\mathcal{F}_E$ = $f'_E $ $ (\text{Node   Mixup} $ $(f(\mathcal{G})))$. The extracted representation $\mathcal{F}_E$ is then passed through a projector to generate the corresponding predictions. These predictions are then compared against the ground-truth labels using the loss function, yielding the environment feature loss defined as:
\begin{equation}
\small
\mathcal{L}_{E}
=
-\frac{1}{n}
\sum_{i=1}^{n}
\mathbf{y}_i^{\top}
\log \left(
{\rm \texttt{Proj}}\left(
f'_{\theta_E}\left(
{\rm \texttt{NodeMixup}}\left(
f(\mathcal{G})
\right)
\right)_i
\right)
\right).
\label{eq:environment_feature_loss}
\end{equation}

Additionally, the extracted environment feature $\mathcal{F}_E$ is utilized to compute the MSE distance between the local representation $R_{local}$ (encodes the structural information of a target node and its one-hop neighborhood) and the global representation $R_{global}$ (captures the node's topological characteristics within the entire graph).
\begin{equation}
    \mathcal{L}_{DS} = \texttt{DS}(\mathcal{F}_E, R_{local}) + \texttt{DS}(\mathcal{F}_E, R_{global}).
\end{equation}
Here, \(DS(\cdot, \cdot)\) denotes the function used to compute the MSE distance. The underlying intuition behind the objective function is to encourage the extracted environment features to diverge from both the local structural patterns of individual classes and the global structural characteristics of the entire graph. This design aims to mitigate the impact of topological imbalance by promoting structural independence in the generated environmental features. The overall training objective of the environmental feature extractor can be formulated as:
\begin{equation}
    \mathcal{L}_{\text{EFE}} = -\mathcal{L}_{E} + d \mathcal{L}_{DS}.
\end{equation}

It is important to note that $d$ is a hyperparameter balancing the two
objectives in $\mathcal{L}_{\mathrm{EFE}}$. 
$-\mathcal{L}_{E}$ encourages the environment extractor to learn
features by reducing their class-discriminative
capacity. In contrast, $d\mathcal{L}_{DS}$ constrains the extracted
environment features to remain structurally consistent with the local and
global representations, preventing arbitrary perturbations. During the
environment-update step, we minimize $\mathcal{L}_{\mathrm{EFE}}$ with
respect to the environment feature extractor, while the remaining modules are not updated.

\subsection{Gated Augmentor}

To further enhance the model’s discriminative capability, GraphIFE incorporates a data augmentation strategy. Specifically, the augmentor generates an augmented feature representation, denoted as $\mathcal{F}_m$, by proportionally mixing invariant features with environment features. This combination enables the model to learn from a richer set of representations, thereby improving its ability to generalize across diverse and imbalanced graph structures. However, determining the optimal proportion for mixing invariant and environment features is non-trivial. To address this challenge, GraphIFE employs a gated augmentor module, denoted as $f_G$, which is designed to learn the appropriate mixing ratio. 

The gated augmentor introduces controlled noise (in the form of environment features) into the invariant features to perturb the model’s discriminative capability and encourage generalization. Specifically, the augmentor incorporates a learnable linear layer that takes the environment features as input and produces a gating value $g$. This gating value is then passed through a sigmoid activation function $\sigma$ to compute the mixing proportion. Finally, the gated augmentor generates the augmented feature $\mathcal{F}_m$ by combining the invariant and environment features according to the learned proportion. The overall mechanism of the Gated Augmentor can be formally summarized as follows:
\begin{equation}
    g = \sigma (f_G(F_I, F_E)), \quad \mathcal{F}_m = F_I + g \odot F_E.
\end{equation}
Here, $\mathbf{g}\in[0,1]^{n'\times d}$ is a dimension-wise gate, and
$\odot$ denotes element-wise multiplication. In the aforementioned data augmentation process, the model autonomously determines the extent to which environment features should be integrated into each feature dimension. During this process, the model dynamically learns dimension-specific weighting coefficients that govern the integration of environment features into each feature representation. By doing so, the augmentor can mitigate issues such as non-convergence of the loss function or the dominance of environment features that may result from direct, unregulated feature superposition. Subsequently, the data augmentation loss is computed to guide the model in learning robust and discriminative feature representations:
\begin{equation}
    \mathcal{L}_{\text{DA}} = -\frac{1}{n} \sum_{i=1}^n y_i^\text{T} \log {\rm \texttt{Proj}}(F_I + \sigma (f_G(F_I, F_E)) F_E)_i),
\end{equation}

\textbf{Regulation.} During training, the gate values may become excessively large, causing the environment features to dominate the representation and introduce excessive noise. Conversely, if the gate values are too small, the environment features may be effectively ignored, rendering the augmentation ineffective and limiting the model’s ability to leverage environment information. To mitigate excessive influence from environment features, we introduce a regularization term that encourages the gate values to remain close to zero, thereby minimizing unnecessary interference. Specifically, this regularization is implemented as an $\ell_1$ sparsity constraint, denoted as $g^\prime$.

Given that the invariant feature extractor must simultaneously optimize multiple loss functions for effective convergence, we employ Dynamic Weight Averaging (DWA) \cite{liu2019end} to adaptively generate the weighting vector  $W_D$ = $[w_{d1}, w_{d2}]$. This approach enables balanced training and promotes more stable and efficient model convergence. The overall training objective for the invariant feature extractor can be formally expressed as follows:
\begin{equation}
    L_{IFE}  = w_{d1} \mathcal{L}_{IF} + w_{d2} \mathcal{L}_{DA} + \alpha g^\prime,
\end{equation}
where $w_{d1}$  and $w_{d2}$ denote the adaptive weights generated by DWA, and $\alpha$ is a hyperparameter that controls the influence of the regularization term $g^\prime$.

\subsection{Analysis}

In this section, we illustrate how GraphIFE mitigates the quality inconsistency of the synthesized node from a theoretical perspective. To mitigate the quality inconsistency issue of the synthesized node, GraphIFE would extract the invariant features of nodes, including the original training set and the synthesized node. GraphIFE doesn't change the quantity relationship between the original and the synthesized nodes. It decouples the quality and quantity of nodes and mitigates the feature inconsistency issue by making the node representation of intra-class as consistent as possible. Although \(P(X_{\mathrm{syn}} \mid Y_{\min})\) may differ from 
\(P(X_{\mathrm{ori}} \mid Y_{\min})\), GraphIFE aims to reduce the discrepancy between
\(P(R_{\mathrm{syn}} \mid Y_{\min})\) and 
\(P(R_{\mathrm{ori}} \mid Y_{\min})\) in the latent space, where \(R\) denotes the learned representation.

In summary, GraphIFE would affect the representation of nodes as follows:  Given $v \in V_{ori}$, $\bar{v}$ $\in V_{syn}$. Let $m_x$ = \texttt{AGG} $\left(\left\{h_u\mid u\in\mathcal{N}(x)\right\}\right)$ denote the  representation of the neighbors of node $x$. $\phi^{(l)}$ denotes the update function. For the representation of the minority class, we have: 
\begin{equation}
\begin{split}
h_{\min}
&=a\phi\!\left(h_v,m_v\right)
 +(1-a)\phi\!\left(h_{\bar v},m_{\bar v}\right)\\
&\approx\phi\!\left(h_{\bar v},m_{\bar v}\right),
\quad
a=\frac{|\mathcal V_{\mathrm{ori}}|}
{|\mathcal V_{\mathrm{ori}}|+|\mathcal V_{\mathrm{syn}}|}\to0.
\end{split}
\end{equation}

In GraphIFE, it optimizes $m_x^{(l-1)}$, which is the representation of \texttt{AGG} in the message-passing mechanism by NLD weight. It also utilizes different strategies to optimize $h_x$, which is the representation of the node itself. 

To simplify the representation, we set $\Phi(x)$ = $\phi\left(h_x,m_x\right)$. GraphIFE is designed to reduce the discrepancy between the invariant
representations of synthesized and original nodes. Suppose that, after
training, the mean invariant representations of a minority class $c$ satisfy
\begin{equation}
\left\|
\bar{\Phi}_{I,\mathrm{syn}}^{c}
-
\bar{\Phi}_{I,\mathrm{ori}}^{c}
\right\|_2
\leq \epsilon_c,
\label{eq:alignment_condition}
\end{equation}
where $\epsilon_c$ characterizes the remaining synthetic-original
representation discrepancy. The mean invariant representation of the balanced
minority class is
\begin{equation}
\bar{h}_{I,\min}^{c}
=
a_c\bar{\Phi}_{I,\mathrm{ori}}^{c}
+
(1-a_c)\bar{\Phi}_{I,\mathrm{syn}}^{c}.
\end{equation}
It therefore follows that
{\small
\begin{equation}
\left\|
\bar{h}_{I,\min}^{c}
-
\bar{\Phi}_{I,\mathrm{ori}}^{c}
\right\|_2
=
(1-a_c)
\left\|
\bar{\Phi}_{I,\mathrm{syn}}^{c}
-
\bar{\Phi}_{I,\mathrm{ori}}^{c}
\right\|_2
\leq
(1-a_c)\epsilon_c.
\label{eq:minority_representation_bound}
\end{equation}
}

At the model level, the expected loss over the balanced minority class can be expressed as:
{\small
\begin{equation}
\begin{split}
&\mathbb{E}_{(u,y)\sim\mathcal{D}^{c}}
\left[\ell\left(f_I(u),y\right)\right]\\
&=
a_c\mathbb{E}_{(v,y)\sim\mathcal{D}_{\mathrm{ori}}^{c}}
\left[\ell\left(f_I(v),y\right)\right]
+
(1-a_c)\mathbb{E}_{(\bar v,y)\sim\mathcal{D}_{\mathrm{syn}}^{c}}
\left[\ell\left(f_I(\bar v),y\right)\right]\\
&\approx
\mathbb{E}_{(v,y)\sim\mathcal{D}_{\mathrm{ori}}^{c}}
\left[\ell\left(f_I(v),y\right)\right].
\end{split}
\label{eq:minority_expected_risk}
\end{equation}
}

That is, the expected risk over the synthesized minority class is encouraged to remain close to that of the original nodes, thereby reducing the model's sensitivity to quality inconsistency.

\begin{table}[b]
\centering
\caption{Statistics of benchmark datasets.}
\label{table: dataset deatils}
\begin{adjustbox}{max width=\columnwidth}
\begin{tabular}{
    >{\bfseries}l
    >{\raggedleft\arraybackslash}p{1.5cm}
    >{\raggedleft\arraybackslash}p{1.8cm}
    >{\raggedleft\arraybackslash}p{1.8cm}
    >{\raggedleft\arraybackslash}p{1.5cm}
}
\toprule
\rowcolor{gray!25}
Dataset & Nodes & Edges & Features & Classes \\
\midrule
Cora      & 2,078   & 10,556   & 1,433   & 7  \\
CiteSeer  & 3,327   & 9,104    & 3,703   & 6  \\
PubMed    & 19,717  & 88,648   & 500     & 3  \\
Photo     & 7,650   & 119,081  & 745     & 8  \\
Computer  & 13,752  & 245,861  & 767     & 10 \\
CS        & 18,333  & 81,894   & 6,805   & 15 \\
\bottomrule
\end{tabular}
\end{adjustbox}
\end{table}

\newcommand{\res}[2]{#1\scriptsize$\pm$#2}
\newcommand{\bestres}[2]{\textbf{\res{#1}{#2}}}
\newcommand{\secondres}[2]{\underline{\res{#1}{#2}}}

\begin{table*}[h]
\centering
\footnotesize
\setlength{\tabcolsep}{8pt}
\renewcommand\arraystretch{1.1}
\caption{Imbalanced node-classification results under the long-tailed setting ($\rho=100$). We report mean $\pm$ standard error over five runs. The best and second-best results are highlighted in \textbf{bold} and \underline{underlining}, respectively.}
\label{table:citation_networks}
\scalebox{0.65}{
\begin{tabular}{ll|lll|lll|lll}
\hline
\multirow{2}{*}{\textbf{}} & \textbf{Dataset} &
\multicolumn{3}{c|}{\textbf{Cora-LT}} &
\multicolumn{3}{c|}{\textbf{CiteSeer-LT}} &
\multicolumn{3}{c}{\textbf{PubMed-LT}} \\ \cline{2-11}
& $\rho=100$ & Acc. & bAcc. & F1 & Acc. & bAcc. & F1 & Acc. & bAcc. & F1 \\ \hline

\multicolumn{1}{c}{\multirow{12}{*}{\rotatebox{90}{GCN}}} &
Vanilla &
\res{73.60}{0.43} & \res{64.13}{0.71} & \res{63.78}{0.68} &
\res{54.56}{0.41} & \res{47.79}{0.38} & \res{43.17}{0.57} &
\res{69.48}{0.98} & \res{56.54}{0.79} & \res{51.06}{0.73} \\
\multicolumn{1}{c}{} &
Reweight &
\res{72.38}{0.18} & \res{64.86}{0.28} & \res{65.23}{0.29} &
\res{54.60}{0.71} & \res{48.09}{0.81} & \res{43.67}{1.17} &
\res{67.90}{0.73} & \res{55.26}{0.58} & \res{49.92}{0.50} \\
\multicolumn{1}{c}{} &
Upsampling &
\res{71.64}{0.49} & \res{63.63}{0.68} & \res{63.97}{0.85} &
\res{54.62}{0.66} & \res{48.03}{0.68} & \res{43.53}{1.01} &
\res{65.76}{0.15} & \res{53.54}{0.12} & \res{48.26}{0.11} \\
\multicolumn{1}{c}{} &
GraphSMOTE &
\res{71.48}{0.21} & \res{62.94}{0.22} & \res{63.39}{0.17} &
\res{53.80}{0.17} & \res{47.25}{0.19} & \res{42.91}{0.37} &
\res{67.90}{0.24} & \res{55.84}{0.18} & \res{51.66}{0.42} \\
\multicolumn{1}{c}{} &
GraphSHA &
\secondres{79.36}{0.29} & \secondres{73.87}{0.38} & \secondres{75.28}{0.30} &
\res{61.50}{0.40} & \secondres{56.18}{0.34} & \secondres{55.18}{0.49} &
\res{71.24}{0.50} & \res{63.04}{1.32} & \res{61.98}{1.27} \\
\multicolumn{1}{c}{} &
GraphENS &
\res{76.64}{0.40} & \res{71.29}{0.69} & \res{71.01}{0.67} &
\secondres{62.22}{0.48} & \res{56.16}{0.59} & \res{55.09}{0.72} &
\secondres{76.90}{0.30} & \res{71.11}{1.23} & \secondres{72.07}{1.14} \\
\multicolumn{1}{c}{} &
BAT &
\res{75.32}{0.49} & \res{66.57}{0.73} & \res{66.09}{0.88} &
\res{57.46}{0.61} & \res{51.00}{0.66} & \res{47.50}{1.04} &
\res{68.58}{0.50} & \res{56.17}{0.47} & \res{51.55}{0.67} \\
\multicolumn{1}{c}{} &
GNNCL &
\res{70.88}{0.32} & \res{60.20}{0.57} & \res{58.13}{0.70} &
\res{52.70}{0.21} & \res{46.14}{0.21} & \res{40.96}{0.35} &
\res{43.24}{2.42} & \res{35.31}{1.89} & \res{23.77}{3.20} \\
\multicolumn{1}{c}{} &
ReNode &
\res{70.82}{1.04} & \res{63.08}{1.46} & \res{64.95}{1.78} &
\res{53.90}{1.63} & \res{49.94}{1.17} & \res{48.68}{1.42} &
\res{68.76}{1.32} & \res{68.14}{1.50} & \res{66.06}{1.42} \\
\multicolumn{1}{c}{} &
TAM &
\res{70.12}{1.27} & \res{62.29}{1.65} & \res{64.10}{2.18} &
\res{56.40}{1.31} & \res{52.38}{1.04} & \res{51.72}{1.03} &
\res{68.86}{1.68} & \res{66.70}{1.86} & \res{65.74}{1.86} \\
\multicolumn{1}{c}{} &
GraphMuGu &
\bestres{79.64}{0.19} & \res{74.43}{0.33} & \bestres{75.35}{0.25} &
\bestres{66.42}{0.27} & \bestres{60.25}{0.22} & \bestres{58.98}{0.45} &
\res{71.16}{0.56} & \secondres{72.16}{1.77} & \res{69.21}{1.18} \\ \cline{2-11}
\rowcolor{light-gray}
\multicolumn{1}{c}{} &
\textbf{GraphIFE} &
\res{79.32}{0.19} & \bestres{74.62}{0.37} & \res{74.78}{0.33} &
\secondres{64.62}{0.59} & \secondres{58.10}{0.58} & \secondres{56.47}{0.62} &
\bestres{77.28}{0.42} & \bestres{74.28}{0.67} & \bestres{74.36}{0.56} \\

\hline

\multirow{12}{*}{\rotatebox{90}{GAT}} &
Vanilla &
\res{73.54}{0.33} & \res{64.44}{0.53} & \res{64.06}{0.77} &
\res{56.80}{0.32} & \res{50.23}{0.29} & \res{46.65}{0.35} &
\res{70.55}{0.25} & \res{57.40}{0.20} & \res{51.92}{0.18} \\
&
Reweight &
\res{73.62}{0.79} & \res{66.19}{1.19} & \res{66.54}{1.37} &
\res{53.92}{0.22} & \res{47.40}{0.22} & \res{41.91}{0.43} &
\res{63.16}{0.42} & \res{51.45}{0.34} & \res{46.28}{0.34} \\
&
Upsampling &
\res{72.94}{0.67} & \res{65.90}{0.96} & \res{66.19}{1.09} &
\res{54.46}{0.20} & \res{47.92}{0.19} & \res{42.78}{0.29} &
\res{64.10}{0.28} & \res{52.21}{0.23} & \res{47.04}{0.23} \\
&
GraphSMOTE &
\res{73.76}{0.51} & \res{66.23}{0.63} & \res{67.07}{0.92} &
\res{54.90}{0.16} & \res{48.23}{0.16} & \res{43.78}{0.32} &
\res{67.82}{0.43} & \res{55.91}{0.52} & \res{51.96}{0.85} \\
&
GraphSHA &
\secondres{78.18}{0.28} & \secondres{72.67}{0.35} & \bestres{74.06}{0.40} &
\res{61.26}{0.41} & \res{56.05}{0.23} & \res{54.98}{0.63} &
\res{68.82}{0.29} & \res{65.05}{1.06} & \res{64.26}{0.95} \\
&
GraphENS &
\res{77.52}{0.21} & \res{72.18}{0.44} & \res{71.98}{0.65} &
\res{63.76}{0.15} & \res{57.62}{0.22} & \res{56.27}{0.33} &
\secondres{76.10}{0.10} & \res{69.88}{0.39} & \secondres{70.77}{0.44} \\
\multicolumn{1}{c}{} &
BAT &
\res{76.24}{0.17} & \res{67.55}{0.26} & \res{65.69}{0.17} &
\res{57.50}{0.52} & \res{50.70}{0.55} & \res{47.10}{0.82} &
\res{64.52}{1.84} & \res{52.54}{1.48} & \res{47.29}{1.54} \\
\multicolumn{1}{c}{} &
GNNCL &
\res{71.84}{0.36} & \res{61.68}{0.37} & \res{59.76}{0.37} &
\res{54.82}{0.12} & \res{48.16}{0.10} & \res{43.55}{0.16} &
\res{63.28}{0.49} & \res{51.59}{0.43} & \res{46.47}{0.51} \\
\multicolumn{1}{c}{} &
ReNode &
\res{66.98}{0.82} & \res{58.77}{0.77} & \res{60.43}{1.13} &
\res{48.30}{3.34} & \res{44.48}{2.59} & \res{41.73}{2.68} &
\res{66.62}{2.47} & \res{68.29}{2.60} & \res{63.62}{3.15} \\
\multicolumn{1}{c}{} &
TAM &
\res{66.84}{1.15} & \res{58.78}{1.14} & \res{59.75}{1.37} &
\res{54.88}{2.38} & \res{51.21}{1.72} & \res{49.34}{1.90} &
\res{66.94}{3.28} & \res{68.11}{3.45} & \res{64.15}{4.00} \\
\multicolumn{1}{c}{} &
GraphMuGu &
\res{78.04}{0.45} & \res{72.64}{0.42} & \secondres{73.68}{0.32} &
\secondres{64.10}{0.19} & \secondres{58.08}{0.17} & \secondres{56.67}{0.44} &
\res{72.64}{0.82} & \secondres{71.96}{1.31} & \res{70.29}{1.08} \\ \cline{2-11}
\rowcolor{light-gray}
\multicolumn{1}{c}{} &
\textbf{GraphIFE} &
\bestres{78.40}{0.28} & \bestres{73.14}{0.34} & \res{73.28}{0.53} &
\bestres{64.66}{1.10} & \bestres{58.62}{1.07} & \bestres{57.05}{1.42} &
\bestres{77.14}{0.43} & \bestres{73.08}{0.83} & \bestres{73.57}{0.73} \\

\hline

\multirow{12}{*}{\rotatebox{90}{SAGE}} &
Vanilla &
\res{71.56}{0.29} & \res{60.27}{0.36} & \res{60.57}{0.29} &
\res{50.30}{0.40} & \res{44.00}{0.35} & \res{39.34}{0.55} &
\res{63.50}{1.10} & \res{51.72}{0.88} & \res{46.45}{0.92} \\
\multicolumn{1}{c}{} &
Reweight &
\res{72.66}{0.57} & \res{63.23}{0.85} & \res{63.77}{0.86} &
\res{54.88}{0.33} & \res{48.26}{0.32} & \res{43.47}{0.34} &
\res{69.80}{1.64} & \res{57.79}{1.63} & \res{54.10}{2.07} \\
\multicolumn{1}{c}{} &
Upsampling &
\res{71.62}{0.66} & \res{62.24}{0.85} & \res{62.76}{0.95} &
\res{54.98}{0.25} & \res{48.44}{0.17} & \res{44.04}{0.19} &
\res{73.44}{0.93} & \res{63.20}{1.72} & \res{62.04}{2.31} \\
&
GraphSMOTE &
\res{73.32}{0.84} & \res{64.11}{0.96} & \res{64.86}{0.91} &
\res{54.02}{0.35} & \res{47.78}{0.37} & \res{44.56}{0.55} &
\res{73.60}{1.46} & \res{63.85}{1.85} & \res{63.66}{2.41} \\
&
GraphSHA &
\secondres{78.60}{0.18} & \bestres{73.20}{0.29} & \bestres{74.24}{0.23} &
\res{60.38}{0.69} & \res{56.90}{0.60} & \secondres{56.73}{0.56} &
\res{66.32}{0.50} & \res{63.18}{1.29} & \res{61.64}{1.08} \\
&
GraphENS &
\res{76.46}{0.29} & \res{70.30}{0.37} & \res{69.83}{0.39} &
\secondres{63.58}{0.33} & \res{57.12}{0.32} & \res{55.21}{0.19} &
\secondres{77.80}{1.20} & \res{71.21}{2.06} & \res{72.35}{2.20} \\
\multicolumn{1}{c}{} &
BAT &
\res{76.68}{0.53} & \res{69.72}{0.73} & \res{70.39}{0.84} &
\res{56.70}{0.54} & \res{50.81}{0.59} & \res{46.56}{0.84} &
\res{70.58}{0.42} & \res{57.59}{0.27} & \res{52.50}{0.18} \\
\multicolumn{1}{c}{} &
GNNCL &
\res{71.84}{0.36} & \res{61.68}{0.37} & \res{59.76}{0.37} &
\res{54.82}{0.12} & \res{48.16}{0.10} & \res{43.55}{0.16} &
\res{63.28}{0.49} & \res{51.59}{0.43} & \res{46.47}{0.51} \\
\multicolumn{1}{c}{} &
ReNode &
\res{65.66}{1.02} & \res{55.09}{1.39} & \res{53.95}{2.01} &
\res{40.48}{2.44} & \res{36.70}{1.82} & \res{33.25}{1.38} &
\res{58.44}{0.64} & \res{60.01}{0.50} & \res{52.38}{0.94} \\
\multicolumn{1}{c}{} &
TAM &
\res{66.08}{0.76} & \res{56.70}{0.98} & \res{56.31}{1.41} &
\res{56.56}{2.36} & \res{52.19}{1.70} & \res{51.53}{1.70} &
\res{58.00}{1.12} & \res{59.46}{0.91} & \res{52.02}{1.56} \\
\multicolumn{1}{c}{} &
GraphMuGu &
\res{78.36}{0.39} & \res{71.47}{0.93} & \res{72.35}{1.07} &
\res{63.06}{0.43} & \secondres{57.62}{0.68} & \bestres{56.95}{0.80} &
\res{74.78}{0.71} & \bestres{74.39}{0.82} & \secondres{72.97}{0.65} \\ \cline{2-11}
\rowcolor{light-gray}
\multicolumn{1}{c}{} &
\textbf{GraphIFE} &
\bestres{78.64}{0.19} & \secondres{73.11}{0.32} & \secondres{73.48}{0.27} &
\bestres{63.96}{0.84} & \bestres{57.80}{0.83} & \res{55.80}{0.97} &
\bestres{77.92}{0.75} & \secondres{73.87}{1.70} & \bestres{74.19}{1.62} \\

\hline
\end{tabular}
}
\end{table*}

\begin{table*}[h]
\centering
\footnotesize
\setlength{\tabcolsep}{8pt}
\renewcommand\arraystretch{1.1}
\caption{Imbalanced node-classification results under the step-imbalanced setting ($\rho=80$). We report mean $\pm$ standard error over five runs. The best and second-best results are highlighted in \textbf{bold} and \underline{underlining}, respectively.}
\label{table:amazon}
\scalebox{0.65}{
\begin{tabular}{ll|lll|lll|lll}
\hline
\multirow{2}{*}{\textbf{}} & \textbf{Dataset} &
\multicolumn{3}{c|}{\textbf{Photo-ST}} &
\multicolumn{3}{c|}{\textbf{Computer-ST}} &
\multicolumn{3}{c}{\textbf{CS-ST}} \\ \cline{2-11}
& $\rho=80$ & Acc. & bAcc. & F1 & Acc. & bAcc. & F1 & Acc. & bAcc. & F1 \\ \hline

\multicolumn{1}{c}{\multirow{12}{*}{\rotatebox{90}{GCN}}} &
Vanilla &
\res{38.85}{0.09} & \res{46.82}{0.09} & \res{31.08}{0.46} &
\res{60.25}{0.45} & \res{48.73}{1.40} & \res{33.19}{0.18} &
\res{36.92}{0.68} & \res{52.03}{0.59} & \res{26.21}{1.24} \\
\multicolumn{1}{c}{} &
Reweight &
\res{44.41}{4.17} & \res{50.12}{2.24} & \res{32.77}{2.86} &
\res{61.87}{0.09} & \res{49.62}{0.54} & \res{33.18}{0.43} &
\res{39.32}{0.84} & \res{51.35}{1.26} & \res{25.47}{2.05} \\
\multicolumn{1}{c}{} &
Upsampling &
\res{42.98}{3.84} & \res{48.96}{1.98} & \res{32.19}{2.15} &
\res{61.77}{0.12} & \res{50.06}{0.56} & \res{33.29}{0.28} &
\res{39.66}{2.25} & \res{52.39}{1.67} & \res{27.80}{2.73} \\
\multicolumn{1}{c}{} &
GraphSMOTE &
\res{45.91}{3.79} & \res{50.40}{1.96} & \res{32.09}{2.41} &
\res{61.83}{0.09} & \res{50.47}{0.38} & \res{33.78}{0.40} &
\res{39.86}{1.37} & \res{52.23}{0.85} & \res{26.56}{1.28} \\
\multicolumn{1}{c}{} &
GraphSHA &
\secondres{83.22}{0.27} & \res{83.05}{0.09} & \res{78.52}{0.26} &
\secondres{75.66}{0.18} & \secondres{83.04}{0.15} & \bestres{71.94}{0.18} &
\secondres{87.14}{0.10} & \secondres{87.63}{0.15} & \secondres{76.34}{0.17} \\
\multicolumn{1}{c}{} &
GraphENS &
\res{81.17}{0.38} & \secondres{83.40}{0.42} & \res{78.64}{0.90} &
\res{75.36}{0.27} & \res{82.72}{0.16} & \res{69.72}{0.35} &
\res{85.05}{0.54} & \res{85.45}{0.30} & \res{74.68}{0.34} \\
\multicolumn{1}{c}{} &
BAT &
\res{38.47}{0.19} & \res{46.41}{0.11} & \res{30.34}{0.50} &
\res{61.15}{0.09} & \res{45.64}{0.06} & \res{26.82}{0.05} &
\res{32.53}{0.07} & \res{47.84}{0.24} & \res{18.74}{0.39} \\
\multicolumn{1}{c}{} &
GNNCL &
\res{38.68}{0.31} & \res{46.63}{0.32} & \res{29.07}{1.10} &
\res{49.74}{7.97} & \res{37.11}{5.45} & \res{22.60}{4.13} &
\res{31.83}{0.21} & \res{46.73}{0.24} & \res{19.24}{0.99} \\
\multicolumn{1}{c}{} &
ReNode &
\res{71.20}{0.27} & \res{77.03}{0.23} & \res{67.72}{0.33} &
\res{74.56}{0.33} & \res{82.25}{0.15} & \secondres{70.70}{0.71} &
\res{86.21}{0.35} & \res{86.64}{0.13} & \res{75.57}{0.19} \\
\multicolumn{1}{c}{} &
TAM &
\res{66.96}{0.39} & \res{72.04}{0.39} & \res{63.41}{0.44} &
\res{68.85}{0.44} & \res{74.96}{0.55} & \res{63.62}{1.10} &
\res{84.16}{0.55} & \res{84.54}{0.17} & \res{72.49}{0.24} \\
\multicolumn{1}{c}{} &
GraphMuGu &
\res{82.81}{0.78} & \res{82.99}{0.55} & \secondres{79.17}{0.80} &
\res{74.50}{0.32} & \res{81.44}{0.37} & \res{69.53}{0.71} &
\res{86.16}{0.59} & \bestres{88.12}{0.16} & \res{73.61}{1.25} \\ \cline{2-11}
\rowcolor{light-gray}
\multicolumn{1}{c}{} &
\textbf{GraphIFE} &
\bestres{83.42}{0.28} & \bestres{86.15}{0.21} & \bestres{81.83}{0.26} &
\bestres{76.38}{0.43} & \bestres{83.99}{0.25} & \res{70.24}{0.75} &
\bestres{87.24}{0.22} & \res{86.68}{0.26} & \bestres{76.55}{0.39} \\

\hline

\multirow{12}{*}{\rotatebox{90}{GAT}} &
Vanilla &
\res{39.55}{0.04} & \res{47.54}{0.04} & \res{25.33}{0.09} &
\res{62.66}{0.31} & \res{48.56}{1.88} & \res{32.62}{3.11} &
\res{43.66}{2.82} & \res{58.14}{2.47} & \res{34.99}{4.01} \\
&
Reweight &
\res{61.78}{0.50} & \res{60.73}{0.09} & \res{47.89}{0.26} &
\res{62.92}{0.05} & \res{46.80}{0.06} & \res{29.72}{0.11} &
\res{43.62}{1.52} & \res{54.44}{1.28} & \res{30.40}{2.01} \\
&
Upsampling &
\res{62.32}{0.24} & \res{61.35}{0.31} & \res{49.40}{0.78} &
\res{62.84}{0.02} & \res{46.76}{0.03} & \res{29.64}{0.07} &
\res{40.16}{2.06} & \res{53.12}{1.45} & \res{27.11}{2.00} \\
&
GraphSMOTE &
\res{62.41}{0.24} & \res{61.53}{0.65} & \res{49.95}{1.58} &
\res{62.84}{0.02} & \res{46.87}{0.02} & \res{29.64}{0.10} &
\res{41.37}{2.62} & \res{54.89}{1.57} & \res{30.13}{2.59} \\
&
GraphSHA &
\res{71.22}{2.89} & \res{75.60}{2.68} & \res{68.23}{2.64} &
\res{70.78}{1.59} & \res{76.70}{0.87} & \res{62.74}{1.85} &
\res{82.25}{1.12} & \res{82.94}{0.58} & \res{65.20}{0.92} \\
&
GraphENS &
\secondres{83.45}{0.20} & \bestres{85.93}{0.38} & \bestres{82.08}{0.29} &
\secondres{76.20}{0.50} & \secondres{83.63}{0.34} & \secondres{71.26}{0.65} &
\secondres{87.54}{0.31} & \bestres{87.81}{0.28} & \bestres{76.17}{1.41} \\
\multicolumn{1}{c}{} &
BAT &
\res{39.03}{0.11} & \res{47.08}{0.09} & \res{30.78}{0.80} &
\res{62.92}{0.25} & \res{45.98}{0.10} & \res{32.07}{0.59} &
\res{32.30}{0.07} & \res{47.25}{0.13} & \res{20.93}{1.10} \\
\multicolumn{1}{c}{} &
GNNCL &
\res{39.87}{0.71} & \res{47.90}{0.77} & \res{30.73}{0.70} &
\res{60.95}{1.85} & \res{42.37}{4.25} & \res{28.53}{1.96} &
\res{32.79}{0.36} & \res{47.74}{0.75} & \res{18.66}{1.02} \\
\multicolumn{1}{c}{} &
ReNode &
\res{66.96}{1.33} & \res{66.75}{1.27} & \res{63.90}{1.10} &
\res{64.36}{0.82} & \res{66.08}{0.70} & \res{55.52}{1.67} &
\res{85.53}{0.73} & \res{81.25}{0.71} & \res{66.17}{1.16} \\
\multicolumn{1}{c}{} &
TAM &
\res{58.39}{1.72} & \res{61.11}{1.89} & \res{53.34}{1.76} &
\res{64.69}{0.44} & \res{64.43}{1.99} & \res{54.88}{1.82} &
\res{81.20}{1.80} & \res{76.62}{0.98} & \res{62.75}{0.72} \\
\multicolumn{1}{c}{} &
GraphMuGu &
\res{72.71}{5.67} & \res{74.26}{3.56} & \res{69.52}{4.17} &
\res{72.57}{0.91} & \res{79.25}{0.37} & \res{65.60}{1.27} &
\res{81.69}{0.74} & \res{82.84}{0.63} & \res{65.37}{0.96} \\ \cline{2-11}
\rowcolor{light-gray}
\multicolumn{1}{c}{} &
\textbf{GraphIFE} &
\bestres{83.53}{0.18} & \secondres{84.41}{0.06} & \secondres{81.71}{0.14} &
\bestres{76.62}{0.84} & \bestres{84.08}{0.28} & \bestres{71.55}{0.78} &
\bestres{88.30}{0.68} & \secondres{87.16}{0.81} & \secondres{74.41}{2.30} \\

\hline

\multirow{12}{*}{\rotatebox{90}{SAGE}} &
Vanilla &
\res{43.37}{1.51} & \res{49.62}{0.78} & \res{29.58}{1.35} &
\res{63.46}{0.19} & \res{46.89}{0.11} & \res{29.91}{0.17} &
\res{43.49}{1.04} & \res{56.44}{0.66} & \res{32.04}{1.15} \\
&
Reweight &
\res{54.31}{2.35} & \res{56.52}{0.95} & \res{41.88}{1.75} &
\res{61.47}{0.62} & \res{49.12}{1.32} & \res{34.27}{0.20} &
\res{53.18}{1.39} & \res{61.52}{0.78} & \res{40.91}{1.43} \\
&
Upsampling &
\res{56.29}{1.16} & \res{57.67}{1.20} & \res{43.97}{1.70} &
\res{62.80}{0.03} & \res{52.40}{1.90} & \res{37.24}{2.38} &
\res{51.99}{1.99} & \res{62.63}{0.64} & \res{41.62}{1.13} \\
&
GraphSMOTE &
\res{56.68}{1.27} & \res{56.24}{0.54} & \res{41.52}{1.04} &
\res{61.35}{0.45} & \res{48.73}{1.40} & \res{33.19}{1.80} &
\res{53.08}{1.72} & \res{63.64}{0.88} & \res{40.93}{1.19} \\
&
GraphSHA &
\secondres{83.93}{0.13} & \secondres{84.93}{0.17} & \secondres{79.88}{0.33} &
\secondres{76.86}{0.17} & \secondres{84.98}{0.10} & \secondres{71.94}{0.18} &
\res{87.27}{0.11} & \secondres{88.17}{0.14} & \res{73.54}{1.23} \\
&
GraphENS &
\res{81.87}{0.38} & \res{83.40}{0.42} & \secondres{78.64}{0.90} &
\res{72.51}{0.35} & \res{79.16}{0.71} & \res{65.44}{0.72} &
\res{85.61}{0.55} & \res{86.53}{0.24} & \secondres{75.05}{0.74} \\
\multicolumn{1}{c}{} &
BAT &
\res{45.99}{2.04} & \res{53.04}{1.22} & \res{35.04}{2.15} &
\res{63.01}{0.60} & \res{46.83}{0.29} & \res{30.43}{0.22} &
\res{63.01}{0.60} & \res{46.83}{0.29} & \res{30.43}{0.22} \\
\multicolumn{1}{c}{} &
GNNCL &
\res{39.44}{0.09} & \res{47.10}{0.10} & \res{28.56}{0.78} &
\res{60.10}{1.21} & \res{44.82}{0.91} & \res{26.83}{0.54} &
\res{39.48}{1.23} & \res{52.97}{0.52} & \res{28.06}{1.06} \\
\multicolumn{1}{c}{} &
ReNode &
\res{56.82}{0.86} & \res{64.25}{0.97} & \res{49.47}{0.64} &
\res{69.70}{0.72} & \res{74.90}{0.13} & \res{62.89}{0.77} &
\res{79.83}{0.24} & \res{81.83}{0.19} & \res{68.60}{0.12} \\
\multicolumn{1}{c}{} &
TAM &
\res{54.33}{0.69} & \res{62.02}{0.77} & \res{46.87}{0.72} &
\res{66.18}{0.60} & \res{66.64}{0.63} & \res{51.87}{0.76} &
\res{82.11}{0.51} & \res{82.21}{0.17} & \res{69.90}{0.30} \\
\multicolumn{1}{c}{} &
GraphMuGu &
\bestres{84.22}{0.59} & \bestres{87.26}{0.50} & \bestres{81.43}{0.48} &
\bestres{77.59}{1.49} & \bestres{85.28}{0.25} & \bestres{72.34}{1.85} &
\bestres{89.09}{0.45} & \bestres{88.92}{0.13} & \res{74.97}{1.60} \\ \cline{2-11}
\rowcolor{light-gray}
\multicolumn{1}{c}{} &
\textbf{GraphIFE} &
\res{82.87}{0.39} & \secondres{85.10}{0.26} & \secondres{80.93}{0.48} &
\res{74.51}{0.53} & \res{81.17}{1.08} & \res{67.82}{0.74} &
\secondres{87.28}{0.60} & \res{87.36}{0.53} & \bestres{76.45}{0.90} \\

\hline
\end{tabular}
}
\end{table*}

\section{Experiments} \label{sec: experiment}

In this section, we present an empirical evaluation of GraphIFE's performance on class-imbalanced node classification tasks. The code is publicly available at \href{https://github.com/flzeng1/GraphIFE}{https://github.com/flzeng1/GraphIFE}.

\subsection{Datasets and Settings}

\textbf{Datasets.} We evaluate the effectiveness of GraphIFE on six benchmark datasets, encompassing both citation networks, Amazon-based co-purchasing, and collaboration graphs. The citation network datasets include (1) Cora, a network in which nodes represent scientific publications and edges denote citation relationships between them; (2) CiteSeer, a similarly structured network with a different scale and statistical properties; and (3) PubMed, a larger citation graph focused on biomedical literature. These datasets are sourced from \cite{sen2008collective}. The Amazon datasets consist of (4) Photo and (5) Computer, both of which are product co-purchasing networks derived from Amazon, where nodes represent products and edges indicate frequently co-purchased item pairs. Additionally, we include the (6) CS dataset, a collaboration network extracted from the Microsoft Academic Graph, where nodes represent authors and edges denote co-authorships \cite{shchur2018pitfalls}. For experimental settings, we follow the long-tail setup for citation networks in \cite{park2021graphens}, and adopt the step-setting configuration for the Amazon datasets, as introduced by \cite{li2023graphsha}. Detailed statistics of all datasets are provided in Table \ref{table: dataset deatils}.

\textbf{Baselines.} We evaluate the performance of GraphIFE by comparing it against baseline methods across three widely used GNN backbones: GCN~\cite{kipf2016semi}, GAT~\cite{velivckovic2017graph}, and GraphSAGE~\cite{hamilton2017inductive}. The comparison includes the following methods: 
(1) Reweight, which applies class-balanced reweighting by adjusting loss weights inversely proportional to class frequencies; 
(2) Upsampling, which directly replicates samples from the minority class to balance the dataset; 
(3) GraphSMOTE ~\cite{zhao2021graphsmote}, which generates synthetic minority nodes by interpolating features of two nodes from the same minority class; 
(4) GraphSHA ~\cite{li2023graphsha}, which employs a novel approach to synthesizing minority nodes by leveraging difficult samples, effectively expanding the decision boundary of minority nodes;
(5) GraphENS~\cite{park2021graphens}, which creates new minority nodes by merging the ego-networks of minority nodes with those of randomly selected nodes; 
(6) BAT~\cite{liuclass}, which addresses topological imbalance by applying dynamic topological augmentation during training. In our experiments, we specifically adopt the BAT0 variant.
(7) GNN-CL \cite{li2024graph} incorporates curriculum learning by adaptively generating interpolated nodes and edges, while jointly optimizing graph classification loss and metric learning. 
(8) ReNode~\cite{chen2021topology}, which reduces the contribution of nodes near class boundaries by down-weighting their loss based on topological proximity; and 
(9) TAM~\cite{song2022tam}, which compares the structural connectivity patterns of each node with a class-averaged counterpart and adaptively adjusts the classification margin accordingly; and 
(10) GraphMuGu \cite{zhao2025generate} jointly synthesizes minority nodes and reweights authentic and generated samples to mitigate class imbalance in graph node classification.
We set the class imbalance ratio to $\rho $ = 100 for the citation network datasets (Cora, PubMed, and CiteSeer), and $\rho$ = 80 for the Amazon datasets (Amazon-Photo, Amazon-Computer, and CS).

\textbf{Evaluation metrics.} We describe the evaluation metrics used to assess model performance across different datasets. Specifically, three metrics are employed: (1) Accuracy (Acc.), which measures the proportion of correctly classified samples relative to the total number of samples; (2) Balanced Accuracy (bAcc.), which accounts for class imbalance by computing the average recall across all classes, thereby providing a more equitable evaluation in imbalanced settings; and (3) F1 Score (F1), defined as the harmonic mean of precision and recall, which offers a balanced assessment of the model’s ability to correctly identify both positive and negative instances. These three metrics are consistently applied to evaluate the performance of all baseline methods across the considered datasets.

\begin{figure}
    \centering
    \includegraphics[scale=0.30]{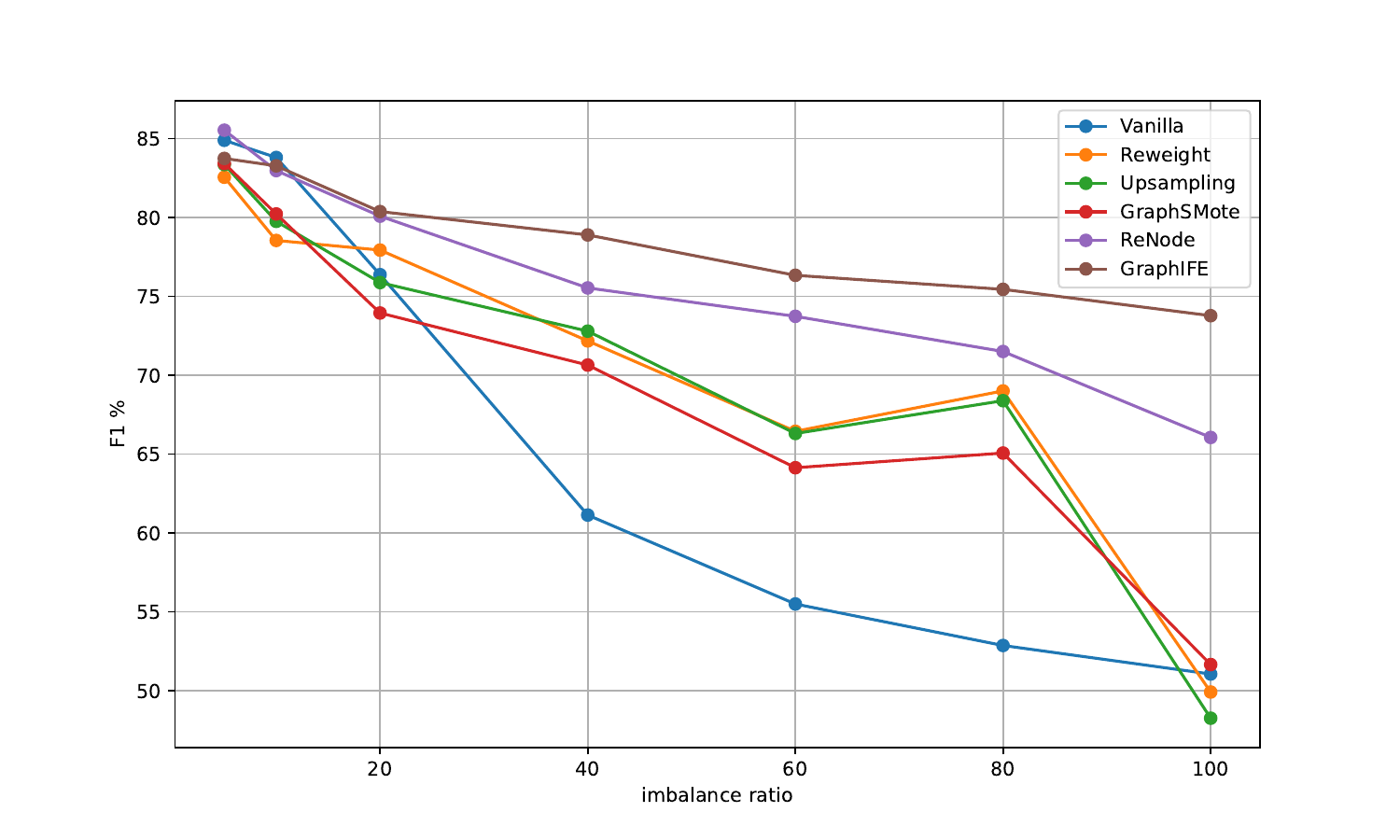}
    \caption{Changing trend of F1-score with the increase of imbalance ratio on PubMed-LT with GCN.}
    \label{fig: imbalance ratio change}
\end{figure}

\subsection{Main Results}
We evaluate the performance of GraphIFE on six different datasets and compared it with various baselines. It should be noted that the citation network datasets (Cora, CiteSeer, and PubMed) follow the LT class imbalance setting \cite{park2021graphens}, while the Amazon datasets (Photo, Computer) and collaboration network datasets (CS) utilize the step class imbalance setting \cite{chen2021topology}. Additionally, the imbalance ratio $\rho$ = 100 in the LT class imbalance setting, while $\rho$ = 80 in the step class imbalance setting. The feature dimension of each GNN is fixed at 256, and the number of attention heads in GAT is set to 8.

 The performance of various baselines compared with GraphIFE on the citation networks is presented in Table \ref{table:citation_networks}. Overall, GraphIFE achieves competitive performance across all reported metrics, datasets, and backbones, demonstrating its robustness under the LT class-imbalance setting. In particular, GraphIFE attains the best results across all three metrics on PubMed-LT with both GCN and GAT and achieves the best accuracy, balanced accuracy, and macro-F1 on CiteSeer-LT with GAT. With GCN, GraphIFE achieves the best balanced accuracy on Cora-LT and remains second on all CiteSeer-LT metrics. With SAGE, it achieves the best accuracy on Cora-LT, the best accuracy and balanced accuracy on CiteSeer-LT, and the best accuracy and macro-F1 on PubMed-LT. GraphMuGu is particularly competitive on Cora-LT and CiteSeer-LT, while GraphIFE retains clear advantages on PubMed-LT across the evaluated backbones. These results indicate that GraphIFE provides robust performance under severe long-tailed class imbalance.

The performance of various baselines compared with GraphIFE on the Amazon datasets and collaboration network datasets is presented in Table \ref{table:amazon}. Under the step class-imbalance setting, GraphIFE also achieves competitive results on every reported metric. With GCN, GraphIFE achieves the best performance on all three metrics for Photo-ST, the best accuracy and balanced accuracy for Computer-ST, and the best accuracy and macro-F1 for CS-ST. With GAT, GraphIFE achieves the best accuracy for Photo-ST and CS-ST, the best performance on all metrics for Computer-ST, and the best or second-best result on all remaining metrics. The results with SAGE are more mixed: GraphMuGu performs best on Photo-ST and Computer-ST, whereas GraphIFE achieves the best macro-F1 and the second-best accuracy on CS-ST. Overall, GraphIFE is particularly effective with GCN and GAT across the step-imbalanced benchmarks, while the comparison also reveals remaining backbone-dependent limitations under SAGE.

\subsection{Results on Large-Scale Graph}

The class imbalance problem is also prevalent in large-scale real-world graphs. To evaluate the generalizability of GraphIFE, we test its performance on a naturally imbalanced large-scale graph. Specifically, we conduct experiments on the OGB-Arxiv dataset \cite{hu2020ogb}, which exhibits severe class imbalance, with imbalance ratios of 775 for the training set, 2,282 for the validation set, 2,148 for the test set, and 942 overall. Given the substantial scale of OGBN-Arxiv, we cap the number of nodes synthesized by GraphIFE at 32,000 per class. We utilize GCN as the backbone of GraphIFE. The baseline results are taken from GraphSHA \cite{li2023graphsha} and the OGB leaderboard \cite{hu2020ogb}. Experimental results are summarized in Table \ref{table:ogbn}, where “OOM” denotes out-of-memory errors. We report validation accuracy, test accuracy, balanced accuracy, and test F1 score. The results show that, except for test bAcc., GraphIFE outperforms all baselines. These findings demonstrate the effectiveness of GraphIFE on large-scale graphs. 

\begin{table}[!t]
\centering
\begin{center}
\caption{Node classification results ($\pm$std) on large-scale naturally class-imbalanced dataset ogbn-arXiv. OOM indicates Out-Of-Memory.}\label{table:ogbn}
\vspace{-0.2cm}
\scalebox{0.80}{
\begin{tabular}{l|cccc}
\toprule
{\textbf{Method}} & {Val Acc.} & {Test Acc.} & {Test bAcc.} & {Test F1} \\
\midrule
Reweight & 67.49\footnotesize{$\pm$0.32} & 66.07\footnotesize{$\pm$0.55} & 53.34\footnotesize{$\pm$0.30} & 48.07\footnotesize{$\pm$0.77} \\
CB Loss & 65.75\footnotesize{$\pm$0.23} & 64.73\footnotesize{$\pm$0.86} & 52.66\footnotesize{$\pm$0.72} & 47.24\footnotesize{$\pm$1.25} \\
Focal Loss & 67.36\footnotesize{$\pm$0.24} & 65.93\footnotesize{$\pm$0.58} & 53.06\footnotesize{$\pm$0.21} & 48.89\footnotesize{$\pm$0.72} \\
ReNode & 66.44\footnotesize{$\pm$0.51} & 65.91\footnotesize{$\pm$0.20} & 53.39\footnotesize{$\pm$0.40} & 48.18\footnotesize{$\pm$0.52} \\
TAM (ReNode) & 67.91\footnotesize{$\pm$0.27} & 66.63\footnotesize{$\pm$0.66} & 
\textbf{53.40\scriptsize$\pm$0.24} & 48.71\footnotesize{$\pm$0.49} \\
GraphSHA & 
73.04\scriptsize$\pm$0.11 & 
72.14\scriptsize$\pm$0.28 &
53.75\scriptsize$\pm$0.16 & 
53.13\scriptsize$\pm$0.20\\
Upsample & 70.53\footnotesize{$\pm$0.08} & 69.55\footnotesize{$\pm$0.37} & 46.82\footnotesize{$\pm$0.07} & 45.49\footnotesize{$\pm$0.20} \\
GraphSMOTE & OOM & OOM & OOM & OOM \\
GraphENS & OOM & OOM & OOM & OOM 
\\
\midrule
\rowcolor{light-gray}
\textbf{GraphIFE} & 
\textbf{73.50\scriptsize$\pm$0.13}& 
\textbf{72.25\scriptsize$\pm$0.30}&
52.05\scriptsize$\pm$0.33& 
\textbf{53.26\scriptsize$\pm$0.39}\\
\bottomrule
\end{tabular}
}
\end{center}
\vspace{-0.5cm}
\end{table}

\begin{figure}
    \centering
    \subfloat[Perspective of varying $d$]{
        \label{fig: d_axis_view}
        \includegraphics[scale = 0.2]{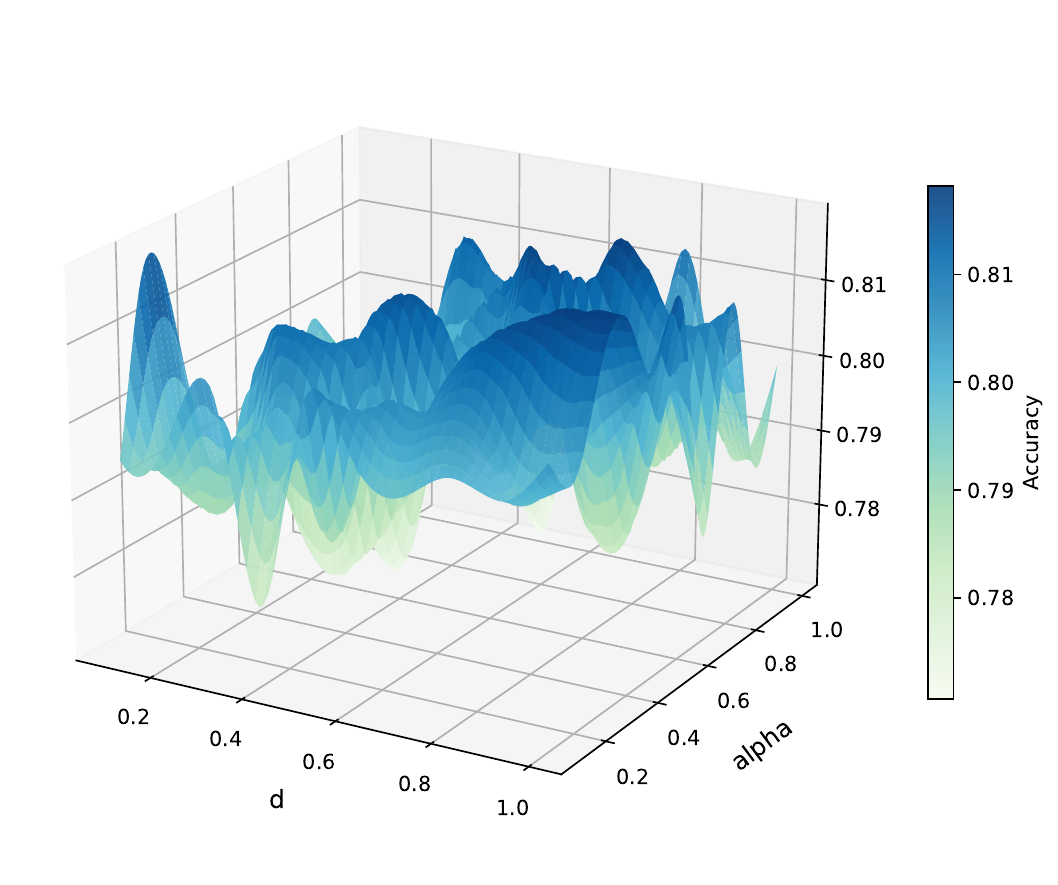}
    }
    
    \subfloat[Perspective of varying $\alpha$]{
        \label{fig: alpha_axis_view}
        \includegraphics[scale = 0.2]{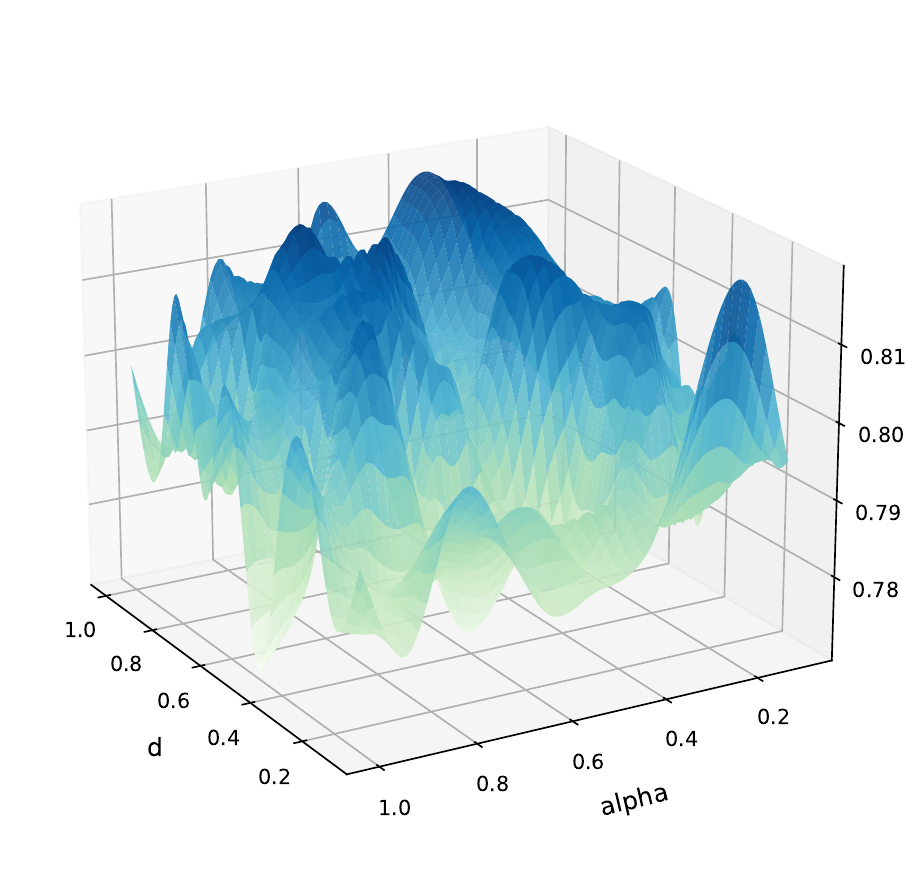}
    }
    \subfloat[Perspective of accuracy]{
        \label{fig: accuracy_view}
        \includegraphics[scale = 0.2]{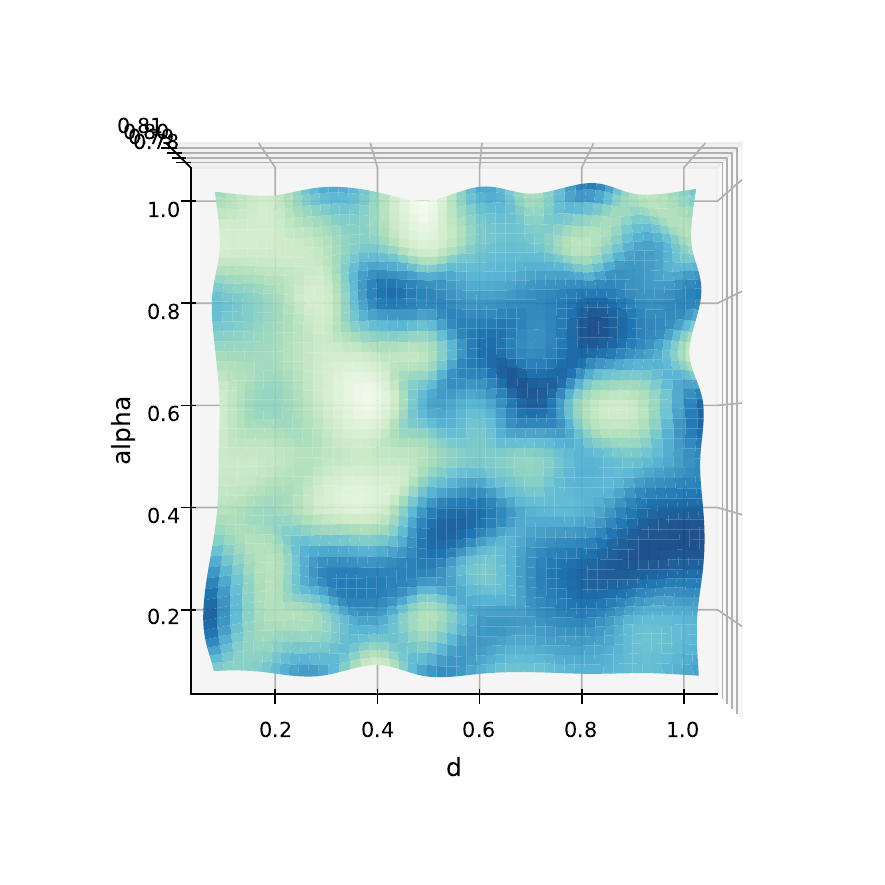}
    }
    \caption{The experimental results of exploring the combined effects of Distance ratio $d$ and Regulation ratio $\alpha$ on the performance of GraphIFE. The experiment is conducted on Photo-ST with GCN as the GNN backbone. Accuracy is utilized as the performance metric.  The color denotes the change in performance. The deeper the color, the better the performance.}
    \label{fig: dis_alpha_combine}
\end{figure}

\subsection{Hyperparameter Analysis}

In this section, we explore the influence of the main hyperparameters on the performance of different datasets. The imbalance ratio $\rho$, the distance ratio $d$, and the regularization ratio $\alpha$ are analyzed in the subsequent section.

\textbf{Imbalance ratio analysis}. We analyze the impact of the class imbalance ratio $\rho$ on model performance, with particular focus on F1-score variations across different methods. The experiment is conducted on the PubMed-LT dataset with GCN as the GNN backbone, and the $\rho$ ranges from 5 to 100. As illustrated in Fig. \ref{fig: imbalance ratio change}, we can observe that F1-scores exhibit statistically significant degradation as $\rho$ increases. While GraphIFE does not achieve peak performance initially, it demonstrates remarkable robustness as $\rho$ increases. GraphIFE remains competitive under increasing imbalance conditions, whereas competing methods show statistically significant F1-score degradation.

\textbf{Distance ratio analysis}. We conduct experiments to analyze the effect of the MSE distance on model performance. The parameter $d$ governs the relative contribution of the MSE distance between local and global representations during environment feature training process. The experiment is conducted on Photo-ST dataset with GCN as a backbone by controlling the value of $d$, systematically examining the impact of parameter $d$. As shown in Fig. \ref{fig: d_axis_view} and Fig. \ref{fig: accuracy_view}, with a fixed $\alpha$, the overall accuracy generally exhibits an upward trend. However, for $\alpha \in \{0.5, 0.6, 0.7\}$, the accuracy first increases, then decreases, and subsequently rises again. In Fig. \ref{fig: accuracy_view}, it is evident that larger values of $d$ lead to better model performance.

\textbf{Regulation ratio analysis}. We conducted the experiment to analyze the regularization term's impact within the invariant feature extractor. $\alpha$ is the hyperparameter that controls the proportion of regularization. The experiment is conducted on the Photo-ST dataset with GCN as the backbone by controlling the value of $\alpha$. As shown in Fig. \ref{fig: alpha_axis_view} and Fig. \ref{fig: accuracy_view}, when $d$ is fixed, the performance does not exhibit a clear linear trend with increasing $\alpha$. Instead, the overall results display a fluctuating pattern, characterized by an initial increase, followed by a decrease, and then a subsequent increase.

In summary, as illustrated in Fig. \ref{fig: accuracy_view}, the closer a region is to the middle or lower-right side of the figure, the darker its color becomes, indicating better performance. This pattern suggests that combining a larger value of $d$ with a moderate value of $\alpha$ yields better overall results.

\begin{figure}
    \centering
    \includegraphics[scale=0.27]{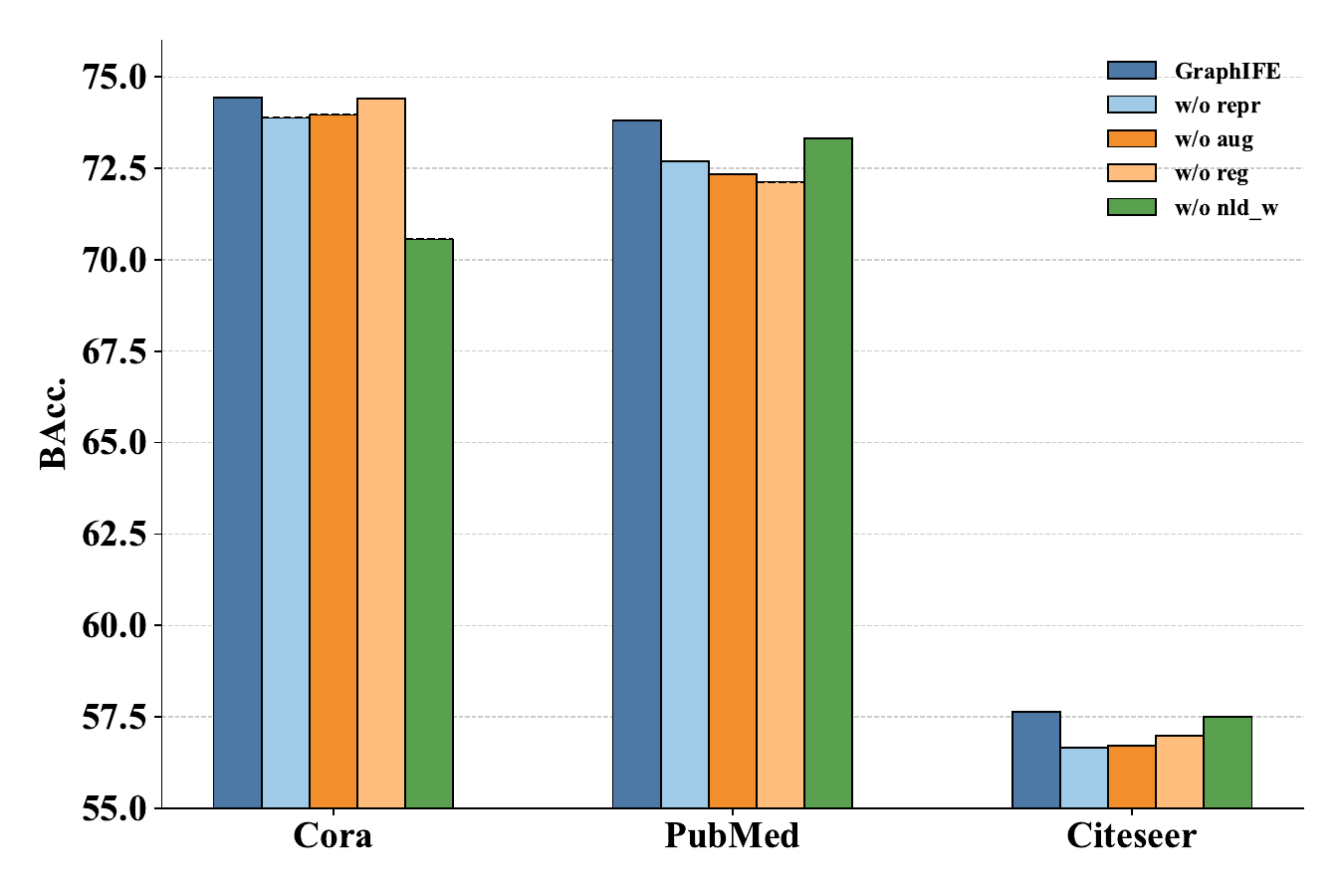}
    \caption{Ablation study of different GraphIFE variants on three datasets.}
    \label{fig: ablation}
\end{figure}

\begin{table}[!t]
\centering
\caption{Classification accuracy for each class on Cora-LT.}\label{table:per}
\scalebox{0.75}{
\begin{tabular}{l|ccccccc}
\toprule
\makecell{\textbf{Class}\\\textbf{Distribution}} &
\makecell{$C_0$\\(0.5\%)} &
\makecell{$C_1$\\(1.1\%)} &
\makecell{$C_2$\\(2.4\%)} &
\makecell{$C_3$\\(5.4\%)} &
\makecell{$C_4$\\(11.6\%)} &
\makecell{$C_5$\\(25.0\%)} &
\makecell{$C_6$\\(54.0\%)} \\

\midrule
Reweight    & 32.9 & 70.9 & 75.0 & 67.5 & 78.6 & 93.5 & 87.9 \\
PC Softmax  & 29.7 & 78.0 & 70.9 & 66.2 & 81.2 & 93.8 & 82.5 \\
CB Loss     & 31.3 & 74.7 & 72.8 & 69.2 & 81.9 & 94.0 & 83.4 \\
Focal Loss  & 29.9 & 75.9 & 72.2 & 71.9 & 82.6 & 94.6 & 83.2 \\
ReNode      & 35.9 & 73.6 & 72.8 & 66.9 & 83.9 & 95.1 & 88.1 \\
\midrule
Upsample    & 12.5 & 58.2 & 65.1 & 67.5 & 76.5 & 92.4 & 89.7 \\
GraphSMOTE  & 22.2 & 66.4 & 68.9 & 62.1 & 79.4 & 93.4 & 89.4 \\
GraphENS    & 37.7 & 72.2 & 73.2 & 63.5 & 74.6 & 94.7 & 82.3 \\
TAM (G-ENS) & 32.6 & 76.3 & 68.9 & 71.5 & 74.2 & 95.1 & 86.9 \\
\rowcolor{light-gray} \textbf{GraphIFE}    & 34.3 & 79.1 & 69.9 & 68.4 & 77.8 & 93.7 & 81.8 \\
\bottomrule
\end{tabular}
}
\end{table}

\subsection{Ablation Study}

This subsection presents an ablation study evaluating the individual contributions of GraphIFE's core components. We assess model performance across three established benchmark datasets (Cora-LT, PubMed-LT, and Citeseer-LT) using GCN as the backbone and balanced accuracy as the evaluation metric. GraphIFE is employed as the baseline to facilitate comparative analysis with the component-ablated variants. The following variants of GraphIFE are considered: (1) ``w/o repr'': GraphIFE without representation alignment during the training of the invariant feature extractor; (2)``w/o aug'': GraphIFE without data augmentation during the training of the invariant feature extractor; (3)``w/o reg'': GraphIFE without regulation during the training of the invariant feature extractor; and (4)``w/o nld\_w'': GraphIFE without neighbor label distribution weight $W_{\textit{NLD}}$ in the neighbor sampling process. 
Rather than removing the entire IFE or EFE branch, we conduct
mechanism-oriented ablations. The variants separately assess the representation alignment and regularization within IFE, the environment-guided augmentation through which EFE contributes to IFE training, and the NLD-guided neighbor sampling strategy.

As illustrated in Fig. \ref{fig: ablation}, the experimental results demonstrate that the complete GraphIFE consistently outperforms all ablation variants. The performance declines when a specific component of GraphIFE is eliminated. The ablation study reveals consistent performance degradation when removing individual components from GraphIFE, which demonstrates each component's essential contribution to the framework's overall effectiveness. The impact of different parts on performance varies in different datasets. w/o repr, w/o aug, and w/o reg obtain a notable decline in performance of the model. The impact of $W_{\textit{NLD}}$ on performance varies with different datasets. This is mainly because of the different NLDs of different datasets. In summary, GraphIFE demonstrates superior performance compared to all variants, while other variants have achieved varying degrees of performance degradation.

We further present a case study on per-class accuracy for baseline methods and GraphIFE with a GCN backbone on Cora-LT, as reported in Table \ref{table:per}. The results for other methods are from a previous study \cite{li2023graphsha}. GraphSMOTE yields only marginal improvements for minority classes compared to simple upsampling, indicating that synthesizing nodes within minority classes alone is insufficient to substantially expand the decision boundary. In contrast, GraphENS achieves relatively high accuracy for minority classes; however, this improvement comes at the expense of reduced performance on major classes, suggesting that GraphENS may excessively generate minority nodes. The per-class results show that GraphIFE improves the overall performance balance between minority and majority classes, although it does not achieve the highest accuracy for every individual class. This finding suggests that GraphIFE primarily mitigates the global imbalance-induced performance gap rather than optimizing each class independently. In particular, GraphIFE improves several minority-class results while maintaining competitive performance on majority classes, leading to a more balanced overall classification behavior.

\begin{table}[h]
    \footnotesize
    \centering
    \caption{Complexity comparison of GraphIFE and different models.}
    \label{table:complexity}
\scalebox{0.65}{
\begin{tabular}{l|l|cc|cc|cc}
\hline
\multirow{2}{*}{\textbf{}} & \multirow{2}{*}{\textbf{Dataset}} 
  & \multicolumn{2}{c|}{\textbf{Vanilla}} 
  & \multicolumn{2}{c|}{\textbf{GraphENS}} 
  & \multicolumn{2}{c}{\textbf{GraphIFE}} \\ \cline{3-8}
& & \textbf{Time} & \textbf{Memory} & \textbf{Time} & \textbf{Memory} & \textbf{Time} & \textbf{Memory} \\
\hline
\multirow{6}{*}{\rotatebox{90}{GCN}}
  & Cora     &  00m 35s & 6.67 MB & 01m 00s& 60.14 MB      &  01m 35s & 178.34 MB     \\
  & CiteSeer &  00m 36s & 15.54 MB &01m 11s & 127.84 MB   &  01m 35s & 327.96 MB    \\
  & PubMed   &  01m 02s & 3.28 MB & 05m 38s & 1511.28  MB  &  07m 40s & 4668.59 MB   \\
  & Photo    &  01m 56s & 3.99 MB &02m 24s & 251.09 MB   &  04m 54s & 858.26 MB    \\
  & Computer &  03m 33s & 4.11 MB  &04m 17s & 769.77 MB    &  09m 16s & 2004.93 MB    \\
  & CS       &  02m 43s & 27.74 MB&04m 05s & 1828.91  MB  &  07m 32s & 3404.69 MB   \\ \hline

\multirow{6}{*}{\rotatebox{90}{GAT}}  
  & Cora     &  00m 32s & 6.68 MB  &00m 55s & 60.15  MB  &  01m 57s & 552.56 MB     \\
  & CiteSeer &  00m 33s & 15.55 MB &01m 05s & 127.85  MB  &  02m 07s & 544.51 MB    \\
  & PubMed   &  01m 15s & 3.29 MB & 06m 04s & 1511.30 MB   &  12m 28s & 6377.69 MB   \\
  & Photo    &  02m 16s & 4.00 MB& 02m 41s & 251.10  MB  &  13m 14s & 4224.71 MB    \\
  & Computer &  04m 26s & 4.13 MB & 05m 15s & 769.79 MB   &  28m 14s & 8993.25 MB    \\
  & CS       &  03m 01s & 27.76 MB &04m 28s & 1828.92 MB  &  16m 00s & 5590.39 MB   \\ \hline

\multirow{6}{*}{\rotatebox{90}{SAGE}} 
  & Cora     &  00m 27s & 13.26MB  &00m 53s & 66.14 MB  &  01m 22s & 288.08 MB     \\
  & CiteSeer &  00m 38s & 31.00 MB  &01m 20s & 143.30 MB  &  01m 56s & 621.91 MB    \\
  & PubMed   &  00m 54s & 6.23 MB  & 05m 23s & 1554.23 MB   &  07m 15s & 4673.54 MB   \\
  & Photo    &  01m 46s & 7.90 MB  & 02m 26s & 255.00  MB  &  04m 23s & 1067.68 MB    \\
  & Computer &  03m 31s & 8.11  MB& 04m 49s & 773.77  MB  &  08m 54s & 2486.06 MB    \\
  & CS       &  10m 19s & 55.33 MB &14m 31s & 1857.04 MB  &  07m 38s & 8719.30 MB   \\ \hline
\end{tabular}
}

\end{table}

\begin{figure}
    \centering
        \subfloat[GraphENS]{
        \label{fig: ENS Visual}
        \includegraphics[scale = 0.33]{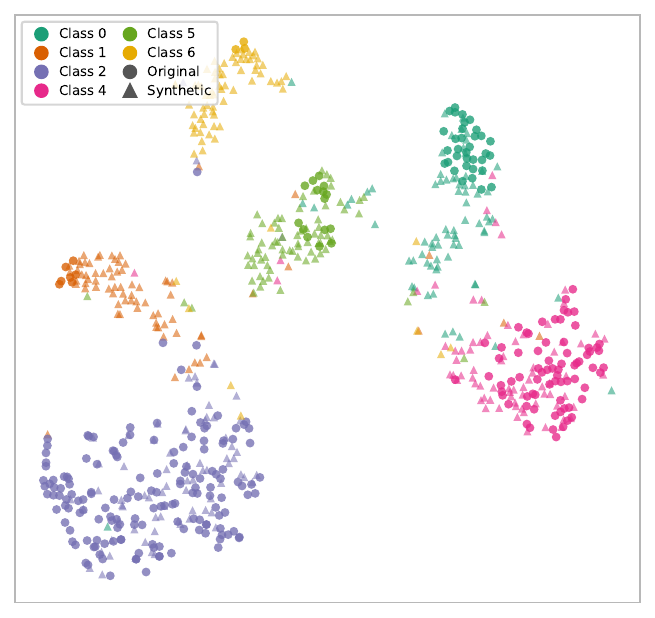}
    }
  
        \subfloat[GraphIFE]{
        \label{fig: IFE Visual}
        \includegraphics[scale = 0.33]{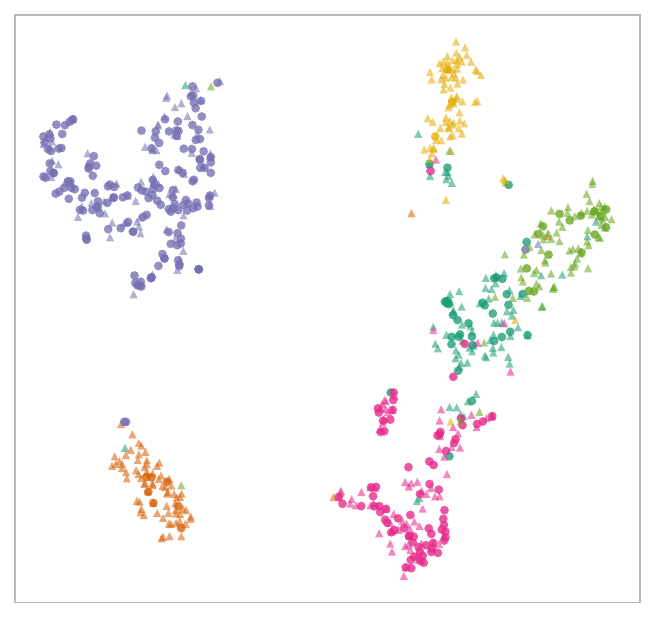}
    }
 \subfloat[GraphSHA]{
        \label{fig: SHA Visual}
        \includegraphics[scale = 0.33]{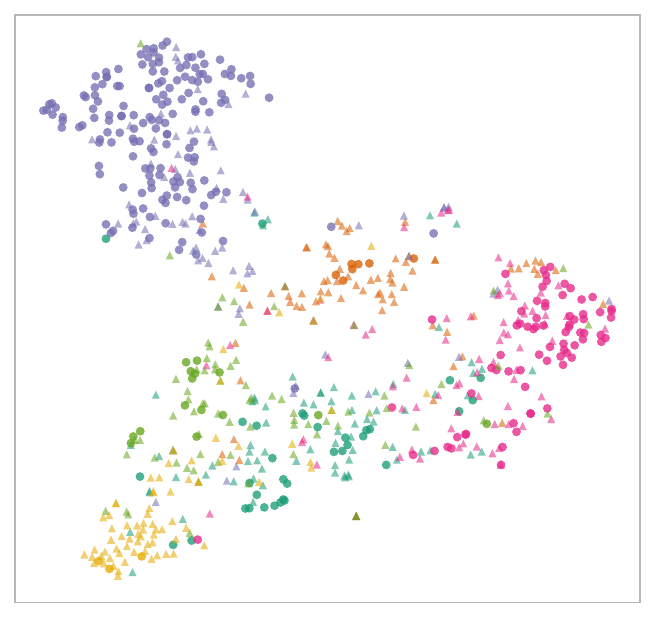}
    }
    \caption{Visualization.}
    \label{fig: Visual}
\end{figure}

\subsection{Complexity Analysis, Visualization, and Limitation}

\textbf{Complexity analysis}. We further analyze the time and peak GPU memory consumption, and then calculate the time complexity of GraphIFE. We compare the complexity of our method with Vanilla and GraphENS to explore its efficiency and scalability. As shown in Table \ref{table:complexity}, GraphIFE consumes more time and memory than Vanilla and GraphENS. Finally, we analyze the time complexity of GraphIFE in the training process. We first denote the numbers of nodes and edges in the dataset as $n$ and $e$. Then, we formally denote the architectural parameters as follows: (1) Layer configurations, $l$, $l_{IF}$, $l_{EF}$ denote the layers of the GNN backbone, invariant feature extractor, and environment feature extractor, respectively; (2) Hidden dimensions: $d$, $d_{IF}$, $d_{EF}$ denote the dimensions of hidden layers in the GNN backbone, invariant feature extractor, and environment feature extractor, respectively; (3) Operational frequencies: $s$ denotes the time of the node synthesis process.  $a$ denotes the time of the gated mixup mechanism. Using the notation above, we first calculate (i) the time complexity of the invariant feature extractor, then (ii) the environment feature extractor, and finally (iii) the complexity of the entire training process. Specifically, to calculate (i), the time complexity of the invariant feature extractor training process includes: the calculation of invariant feature extraction $O((dl + d_{IF}l_{IF})(n + e))$; the calculation of environment feature extraction $O((d_{EF}l_{EF})(n + e))$; the regulation term $O(n)$; and the calculation of representation alignment $O(nd)$. Note that the invariant feature extractor and the environment feature extractor share the same architecture; then $d_{IF}$ = $d_{EF}$, $l_{IF}$ = $l_{EF}$. The time complexity of the invariant feature extractor training process is $O((n + e)(dl + 2d_{IF}l_{IF}) + a + n + s)$. To calculate (ii), the complexity of the environment feature extractor training includes the calculation of environment feature extraction $O((n + e)(dl + dl_{EF}))$ and the calculation of MSE distance $O(2nd)$. Therefore, the time complexity of the environment feature extractor training process is $O((n + e)(dl + d_{EF}l_{EF}) + 2nd)$. Finally, to calculate (iii), the overall time complexity of the full training process is $O((n + e)(2dl + 3d_{EF}l_{EF}) + 2nd + a + n + s)$.

\textbf{Visualization}. As shown in Fig.~\ref{fig: Visual}, we visualize the minority-node representations of GraphENS, GraphSHA, and GraphIFE on Cora-LT using t-SNE. All representations are extracted before the final projector layer; GraphIFE uses the invariant representation $\mathcal{F}_{I}$. Colors denote class labels, while circles and triangles represent original and synthesized nodes, respectively. For visual clarity, we retain all original minority training nodes and randomly sample synthesized nodes per minority class. Compared with GraphENS and GraphSHA, GraphIFE exhibits visually closer original and synthesized representations within most minority classes, with less apparent inter-class intermingling.

\textbf{Limitation}. GraphIFE incurs additional computational and memory costs compared with single-branch rebalancing baselines. Specifically, it maintains separate invariant and environment feature extractors, performs alternating optimization, and constructs synthetic nodes during training. These operations can introduce
non-negligible overhead when applied to large-scale graphs. Additionally, the proposed alignment objectives are empirically motivated and evaluated.

\section{Conclusion and Future Work} \label{sec: Conclusion}

In this paper, we explore the graph imbalance node classification problem and identify a quality inconsistency issue, in which the features of synthetic nodes may be out of distribution during the node synthesis process. To address this issue, we propose an adversarial feature perturbation framework called GraphIFE, which uses an invariant feature extractor to extract the invariant features of both original and synthetic nodes and an environment feature extractor to obtain environment features for data augmentation and the mitigation of topological imbalance. Additionally, GraphIFE employs a gated mechanism to generate mixed features, enhancing the model’s ability to identify invariant node features. Moreover, we propose representation alignment and neighbor label distribution weighting to mitigate the quality inconsistency issue. Overall, GraphIFE demonstrates an effective and robust approach to imbalanced node classification, achieving consistently strong performance across diverse datasets. For future work, we aim to further optimize GraphIFE to reduce computational overhead while maintaining strong performance. Moreover, we expect GraphIFE to generalize to other quantity-based heterogeneous information network which exhibit imbalanced node types.

\bibliographystyle{IEEEtran}
\bibliography{GraphIFE.bib}

\end{document}